%% file: main.tex
\documentclass[10pt,twocolumn,letterpaper]{article}

\usepackage[pagenumbers]{cvpr} 

\usepackage{graphicx}
\usepackage{amsmath}
\usepackage{amssymb}
\usepackage{booktabs}
\usepackage{multirow}
\usepackage[normalem]{ulem}
\usepackage{bm}
\usepackage{algorithm2e}
\usepackage[accsupp]{axessibility}

\usepackage[pagebackref,breaklinks,colorlinks]{hyperref}

\usepackage[capitalize]{cleveref}
\crefname{section}{Sec.}{Secs.}
\Crefname{section}{Section}{Sections}
\Crefname{table}{Table}{Tables}
\crefname{table}{Tab.}{Tabs.}

\def\cvprPaperID{3516} 
\def\confName{CVPR}
\def\confYear{2023}

\begin{document}

\title{Grad-PU: Arbitrary-Scale Point Cloud Upsampling via Gradient Descent with Learned Distance Functions}

\author{Yun He$^{1}$ \quad Danhang Tang$^{2}$ \quad Yinda Zhang$^{2}$ \quad Xiangyang Xue$^{1}$ \quad Yanwei Fu$^{1}$ \\
$^{1}$ Fudan University  \quad $^{2}$ Google
}

\maketitle

\input{0_math.tex}
\input{0_tweaks.tex}
\input{1_abstract.tex}
\input{2_intro.tex}
\input{3_related.tex}
\input{4_method.tex}
\input{5_eval.tex}

\input{6_conclusion.tex}

{\small
\bibliographystyle{ieee_fullname}
\bibliography{main}
}

\clearpage
\input{7_supp.tex}

\end{document}

%% file: 0_math.tex
\newcommand{\point}{p}
\newcommand{\edge}{e}
\newcommand{\stepsize}{\lambda}
\newcommand{\iteration}{T}
\newcommand{\feature}{f}
\newcommand{\dimension}{d}
\newcommand{\upsamplingrate}{R}
\newcommand{\variance}{\sigma^2}
\newcommand{\noiselevel}{\tau}
\newcommand{\distancefunction}{F}

\newcommand{\displacement}{(\delta_{x}, \delta_{y}, \delta_{z})}

\newcommand{\inputpointcloud}{P_L}
\newcommand{\interpolatedpointcloud}{P_I}
\newcommand{\outputpointcloud}{P_H}
\newcommand{\querypointcloud}{P_Q}
\newcommand{\groundtruthpointcloud}{P_G}

\newcommand{\localfeature}{l}
\newcommand{\globalfeature}{g}

%% file: 0_tweaks.tex
\definecolor{turquoise}{cmyk}{0.65,0,0.1,0.3}
\definecolor{purple}{rgb}{0.65,0,0.65}
\definecolor{dark_green}{rgb}{0, 0.5, 0}
\definecolor{green}{rgb}{0, 1.0, 0}
\definecolor{orange}{rgb}{0.8, 0.6, 0.2}
\definecolor{red}{rgb}{0.8, 0.2, 0.2}
\definecolor{blueish}{rgb}{0.0, 0.7, 1}
\definecolor{light_gray}{rgb}{0.7, 0.7, .7}
\definecolor{pink}{rgb}{1, 0, 1}
\definecolor{cyan}{rgb}{0.0, 1.0, 1.0}
\definecolor{dark_orange}{rgb}{1.0, 0.55, 0.0}
\definecolor{sky_blue}{rgb}{0.0, 0.7, 1}

\newcommand{\Alg}[1]{Alg~\ref{alg:#1}}
\newcommand{\Fig}[1]{Fig~\ref{fig:#1}}
\newcommand{\Figure}[1]{Figure~\ref{fig:#1}}
\newcommand{\Tab}[1]{Tab~\ref{tab:#1}}
\newcommand{\Table}[1]{Table~\ref{tab:#1}}
\newcommand{\eq}[1]{(\ref{eq:#1})}
\newcommand{\Eq}[1]{Eq~\ref{eq:#1}}
\newcommand{\Equation}[1]{Equation~\ref{eq:#1}}
\newcommand{\Sec}[1]{Sec~\ref{sec:#1}}
\newcommand{\Section}[1]{Section~\ref{sec:#1}}
\newcommand{\Appendix}[1]{Appendix~\ref{app:#1}}

\renewcommand{\paragraph}[1]{{\vspace{.25em}\noindent \textbf{#1.}}}

\newcommand\blfootnote[1]{%
  \begingroup
  \renewcommand\thefootnote{}\footnote{#1}%
  \addtocounter{footnote}{-1}%
  \endgroup
}

%% file: 1_abstract.tex
\begin{abstract}
    Most existing point cloud upsampling methods have roughly three steps: feature extraction, feature expansion and 3D coordinate prediction.     
     However, they usually suffer from two critical issues:
    (1) fixed upsampling rate after one-time training, since the feature expansion unit is customized for each upsampling rate;
    (2) outliers or shrinkage artifact caused by the difficulty of precisely predicting 3D coordinates or residuals of upsampled points.
    To adress them, we propose a new framework for accurate point cloud upsampling that supports arbitrary upsampling rates. Our method first interpolates the low-res point cloud according to a given upsampling rate. And then refine the positions of the interpolated points with an iterative optimization process, guided by a trained model estimating the difference between the current point cloud and the high-res target.
    Extensive quantitative and qualitative results on benchmarks and downstream tasks demonstrate that our method achieves the state-of-the-art accuracy and efficiency.

\blfootnote{\,\,Yun He and Xiangyang Xue are with the School of Computer Science, Fudan University.} 
\blfootnote{\,\,Yanwei Fu is with the School of Data Science, Fudan University. He is also with Shanghai Key Lab of Intelligent Information Processing, and Fudan ISTBI–ZJNU Algorithm Centre for Brain-inspired Intelligence, Zhejiang Normal University, Jinhua, China.}
    
\end{abstract}

\begin{figure}[t!]
\centering\includegraphics[width=\linewidth]{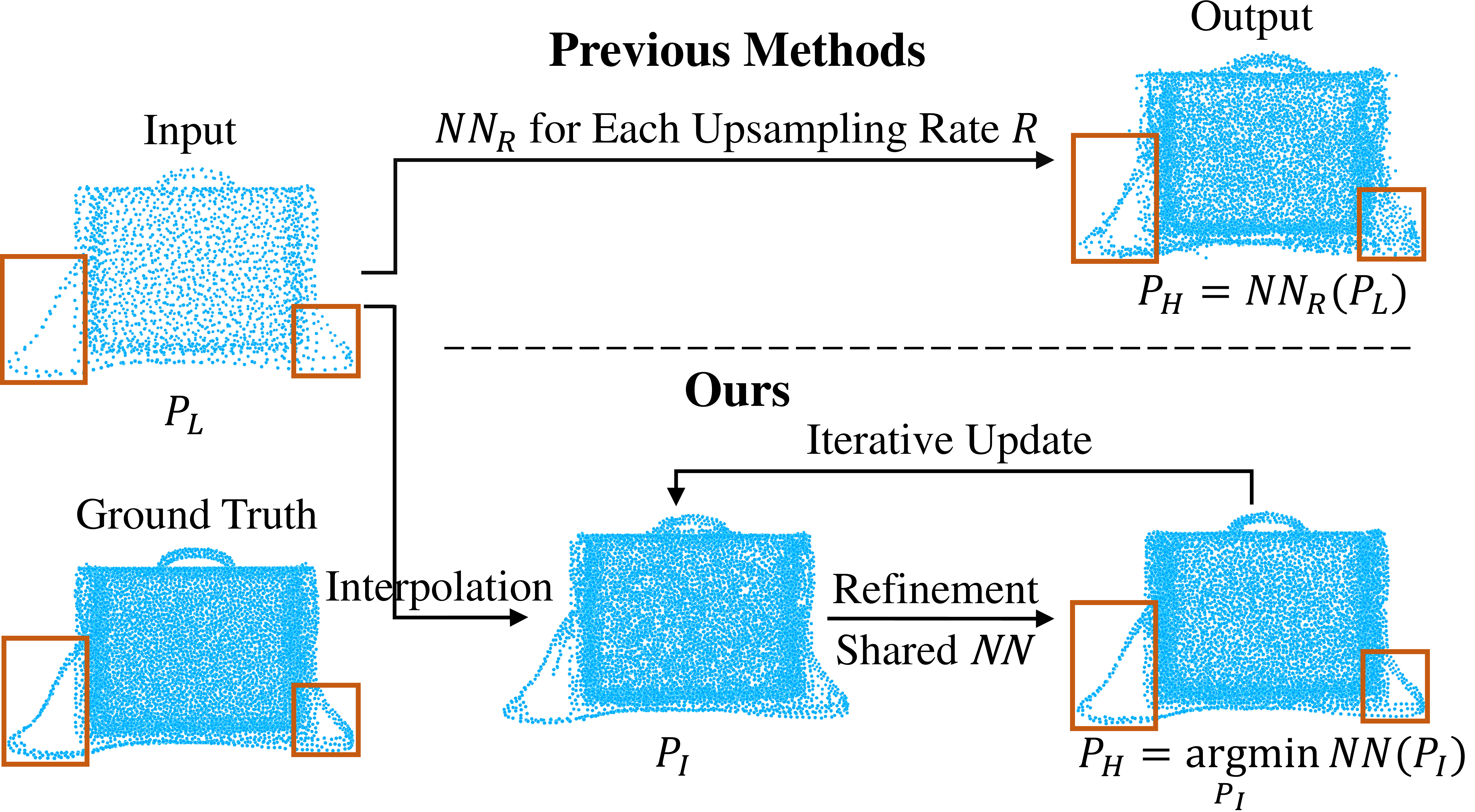}
\vspace{-10pt}
\caption{The comparison between previous point cloud upsampling methods and ours, and $NN$ denotes the deep neural network. 
Given the low-res input $\inputpointcloud$, previous methods directly predict the 3D coordinates or residuals of high-res output $\outputpointcloud$. 
And most of them need retraining to satisfy various upsampling rates. 
Instead we first interpolate points in Euclidean space, which separates point generation from network learning and thus achieves arbitrary upsampling rates.
Then we formulate the refinement of interpolated points as an iterative process aiming
to minimize the learned point-to-point distance function $NN(\interpolatedpointcloud)$.
}
\label{fig:teaser}
\end{figure}

%% file: 2_intro.tex
\section{Introduction}

With the popularity of commercial 3D scanners, capturing point clouds from real-world scenes becomes convenient and affordable. 
Thus point clouds have been widely utilized in applications such as  autonomous driving, robotics, remote sensing, etc \cite{guo2020deep}. That being said, the raw point clouds produced by 3D scanners or depth cameras are often sparse and noisy, sometimes with small holes \cite{li2021point},
which greatly affects the performance of downstream tasks, such as semantic classification \cite{xiang2021walk}, rendering \cite{dai2020neural}, surface reconstruction \cite{bernardini1999ball}, etc.
Consequently, it is vital to 
upsample a raw point cloud to a dense, clean and complete one, with more geometric details.

The common practice towards point cloud upsampling usually consists of three key steps \cite{yu2018pu,yifan2019patch,li2019pu,li2021point,qian2021pu,qiu2021pu,long2022pc2}. (1) Feature extraction: capturing point-wise semantic features from the low-res point clouds. (2) Feature expansion: expanding the extracted features w.r.t the specified upsampling rate.
(3) Coordinate prediction: predicting 3D coordinates or residuals of upsampled points from the expanded features.
However, there are two critical issues in this paradigm.
Firstly, these models are usually dependent on the upsampling rate. To support different upsampling rates, multiple models need to be trained.
Secondly, precisely estimating the 3D coordinates or offsets to the target points is hard, which leads to outliers or shrinkage artifact \cite{luo2021score}.
Although some recent methods try to handle the fixed upsampling rate problem via affine combination of neighboring points \cite{luo2021pu, qian2021deep} or implicit neural representation \cite{feng2022neural, zhao2022self}, their performance is still limited by the inaccuracy of 3D coordinate prediction.

In this paper, we propose a novel point cloud upsampling algorithm to address these two issues. 
In particular, our method decouples the upsampling process into two steps. 
First, we propose to directly upsample the input low-res point cloud in Euclidean space by midpoint interpolation, instead of expanding in the feature space. 
And the amount of interpolated points is determined by a given upsampling ratio.
This makes the learning part independent with
the upsampling module and helps the whole method generalize to arbitrary upsampling rates.
Secondly, the interpolated point cloud is refined by an iterative process aiming to minimize the difference between the interpolated point cloud and the ground truth high-res point cloud. 
To measure the difference, we choose to use point-to-point distance, which eliminates the need of surface extraction and can handle arbitrary topologies. 
Moreover, comparing to coordinates ($\in \mathbb{R}^3$), the point-to-point distance ($\in \mathbb{R}^1$) is an easier objective to optimize, thus results in much more accurate upsampling results in our experiments. 
Since the ground truth point cloud is not available during inference, a model is trained to approximate the point-to-point distance function in a differentiable manner, thus termed as P2PNet. 
To improve the training efficiency, we come up with a simple but effective training scheme, by adding Gaussian noise to the data to simulate varying degrees of difference between the input and ground truth point cloud. 
The P2PNet is then trained to minimize the difference, \ie, the refinement step is regarded as a distance minimization process.

In this paper, we propose a novel framework for accurate point cloud upsampling with arbitrary upsampling rates. 
Specifically, our contributions can be summarized as:
\begin{itemize}
\item Decompose the upsampling problem into midpoint interpolation and location refinement, which achieves arbitrary upsampling rates. 
\item Formulate the refinement step as a point-to-point distance minimization process.
\item Propose the P2PNet to estimate the point-to-point distance in a differentiable way. 
\end{itemize}
Extensive experiments show that our method significantly outperforms existing methods in accuracy, efficiency, robustness, and generalization to arbitrary upsampling rates, also improves the performance of downstream tasks such as semantic classification and surface reconstruction.

%% file: 3_related.tex
\section{Related Work}
\paragraph{Point Cloud Analysis}
Due to the natural irregular structure of point clouds, traditional methods always first voxelize them and then apply 3D convolution for processing \cite{maturana2015voxnet,riegler2017octnet}, which however brings huge computational cost.
Thus some other methods try to operate directly on the raw point clouds \cite{qi2017pointnet,qi2017pointnet++,wang2019dynamic,zhao2021point,fan2021point,fan2022pstnet,li2018pointcnn,wu2019pointconv,he2022density}.
Specifically, PointNet \cite{qi2017pointnet} applies shared MLPs to extract point-wise features first and then uses max pooling to obtain the order-invariant global features.
PointNet++ \cite{qi2017pointnet++} designs the set abstraction operation to further enhance the capture of local geometry.
DGCNN \cite{wang2019dynamic} achieves nonlocal feature diffusion by constructing dynamic graphs in feature space.
Point Transformer \cite{zhao2021point} introduces attention mechanism \cite{vaswani2017attention} to capture the long-range relations.
And Fan \etal \cite{fan2021point} designs the Point 4D Convolution for modeling the spatio-temporal correlations in point cloud sequences.
Considering the effectiveness and efficiency, we simplify it to apply on the spatial domain only, and denote it as Point 3D Convolution.

\paragraph{Learnable Point Cloud Upsampling}
Benefiting from the success of deep learning technology in the point cloud analysis field, researchers begin to focus on the learning-based point cloud upsampling methods \cite{yu2018pu,yu2018ec,yifan2019patch,li2019pu,qian2020pugeo,li2021point,qian2021pu,luo2021pu,qian2021deep,ye2021meta, qiu2021pu,long2022pc2,feng2022neural,zhao2022self}.
In particular, PU-Net \cite{yu2018pu} adopts the PointNet++ \cite{qi2017pointnet++} backbone to first extract multi-level features, then expands them by multi-branch MLPs, and finally transforms the expanded features to 3D coordinates.
MPU \cite{yifan2019patch} proposes the EdgeConv \cite{wang2019dynamic} based feature extractor, and expands features by assigning different 1D codes.
PU-GAN \cite{li2019pu} introduces adversarial training and designs a up-down-up unit for expanded features correction.
PUGeo-Net \cite{qian2020pugeo} first generates points in 2D space and then transforms them to 3D space.
Dis-PU \cite{li2021point} disentangles the upsampling process by a dense generator and spatial refiner.
PU-GCN \cite{qian2021pu} proposes Inception DenseGCN for feature extraction and NodeShuffle for feature expansion.
PU-Transformer \cite{qiu2021pu} introduces a transformer-based model to capture fine-grained point features.
Moreover, PC$^2$-PU \cite{long2022pc2} designs patch correlation and point correlation modules to improve the global spatial consistency.
Besides the extra time-consuming annotations \cite{yu2018ec,qian2020pugeo,yifan2019patch}, these methods usually have two aforementioned issues: fixed upsampling rate after each training and outliers or shrinkage artifact due to the difficulty of 3D coordinate estimation.
Despite a few recent methods break the former limitation by affine combination of neighbor points \cite{luo2021pu,qian2021deep} or implicit function learning \cite{zhao2022self,feng2022neural}, the latter problem still remains unsolved.
To handle these two issues simultaneously,
we propose to upsample points in Euclidean space by midpoint interpolation, and then refine them via distance minimazation.

\paragraph{Implicit Neural Representation}
Learning continuous implicit functions for 3D shape representation has prevailed the research community in recent years \cite{park2019deepsdf,chibane2020neural,chen2019learning,jiang2020local,gropp2020implicit,mescheder2019occupancy,chibane2020implicit}.
Common practice is to train neural networks to approximate conventional implicit shape functions, such as 
occupancy probability\cite{mescheder2019occupancy,chen2019learning,chibane2020implicit}, 
signed distance fields (SDF)~\cite{park2019deepsdf,gropp2020implicit} and unsigned distance fields (UDF)~\cite{chibane2020neural}.

%% file: 4_method.tex
\section{Methodology}
We propose a novel point cloud upsampling framework. Once trained, it can upsample a point cloud with arbitrary ratios. 
Specifically, given a low-res point cloud $\inputpointcloud$, we first interpolate it to obtain a new point cloud $\interpolatedpointcloud$ with desired amount of points in \Sec{interpolation}. Then the locations of the interpolated points are refined by an iterative optimization process to be as close to the ground truth high-res point cloud $\groundtruthpointcloud$ as possible, as in \Sec{refinement}. Since the ground truth is not available during inference, this refinement is guided by a trained model, termed as P2PNet (\Sec{p2pnet}).

\subsection{Midpoint Interpolation}
\label{sec:interpolation}
To make our network learning uncoupled with point generation, thus achieving arbitrary upsampling rates, we propose the midpoint interpolation for point upsampling.
Given the low-res input $\inputpointcloud$, our interpolation method goes through the following two steps.
(1) Midpoint generation: for each point $\point \in \inputpointcloud$, we first find its $k$-nearest neighbor $\point_k$, and then use its midpoint $(\point+\point_k)/2$ as the new generated point.
(2) Farthest point sampling (FPS): to remove repeatedly generated midpoints and control their number w.r.t the desired upsampling rate $\upsamplingrate$,
we apply FPS to downsample the output of previous step. 
And the union of all downsampled points forms the final interpolated result $\interpolatedpointcloud$.

\subsection{Point Location Refinement}
\label{sec:refinement}
The second step is to refine the interpolated point cloud $\interpolatedpointcloud$ to recover the fidelity. We formulate the problem as minimizing the difference between $\interpolatedpointcloud$ and the ground truth point cloud $\groundtruthpointcloud$. To do so, one needs a distance metric.

\subsubsection{Point-to-Point Distance}
\begin{figure}[htbp]
\flushleft
\includegraphics[ width=\linewidth]{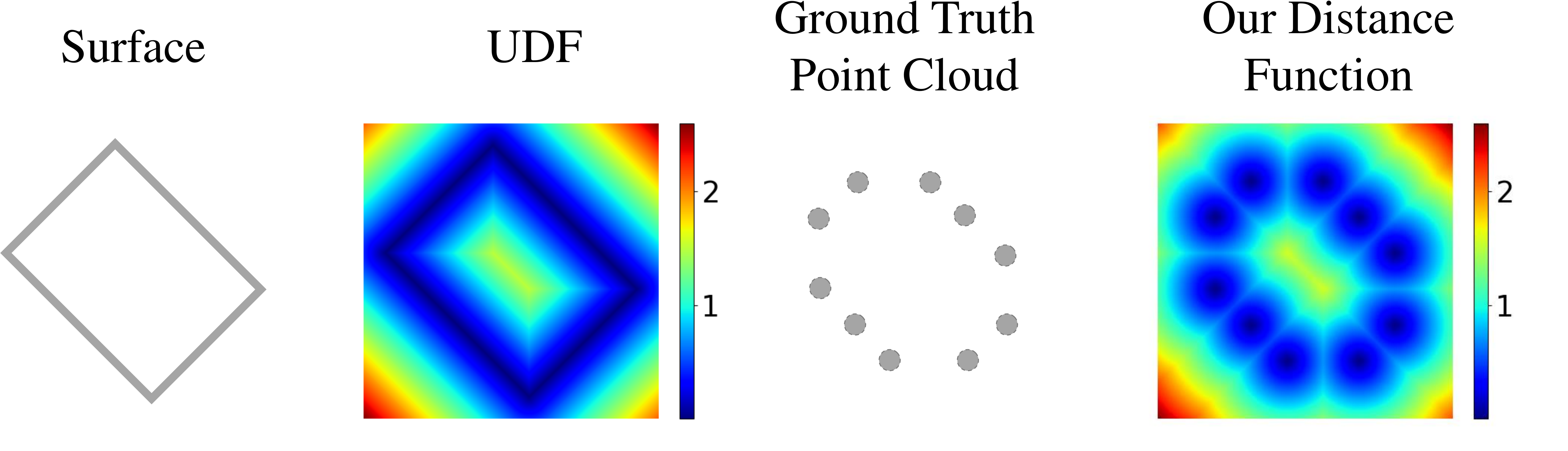}
\caption{2D illustrations of (left) a surface and its UDF; (right) a point cloud and its point-to-point distance function.}
\vspace{-5pt}
\label{fig:df}
\end{figure}

A straightforward metric is to regard the point clouds as implicit surfaces, and measure the differences with the point-to-surface distances, such as SDF \cite{park2019deepsdf,gropp2020implicit} or UDF \cite{chibane2020neural}.
However, it is not always possible to reasonably extract a surface from a low-res point cloud. In contrast, we use the unsigned point-to-point distance function. Specifically, 
given an interpolated point $\point \in \interpolatedpointcloud$, the distance function $\distancefunction(\point)$ represents the Euclidean distance between point $\point$ and its nearest neighbor point $\hat{\point}$ in the ground truth high-res point cloud $\groundtruthpointcloud$. This function does not require a surface and can handle arbitrary
topologies, as \Fig{df} illustrated.

\subsubsection{Distance Minimization}
The location of the newly interpolated points in  $\interpolatedpointcloud$ is naively computed and therefore noisy. To improve the accuracy, they need to be moved towards the ground truth positions. 
A straightforward solution is to predict a coordinate displacement ($\in \mathbb{R}^3$) for each interpolated point $\point \in \interpolatedpointcloud$\cite{yu2018ec,li2021point,luo2021pu}.
However, the prediction is often inaccurate and thus results in outliers or shrinkage artifact \cite{luo2021score}.
To tackle this, we formulate the problem as a distance minimization process. At every iteration, an ``oracle'' will give us the point-to-point distance ($\in \mathbb{R}^1$) between the current point $\point$ and the closest point $\hat{\point}$ in the ground truth high-res point cloud $\groundtruthpointcloud$. Through gradient descent \cite{ruder2016overview}, the distance loss is back-propagated to encourage the interpolated points moving towards the ground truth, as formulated below:
\begin{equation}
\begin{split}
    \point^{t+1}=\point^{t}-\stepsize \nabla \distancefunction(\point^{t}),
    t=0,...,\iteration-1
    \label{eq:update}
\end{split}
\end{equation}
where we have the initial interpolated point $\point^0$, the updated point $\point^{t+1}$, the step size $\stepsize$, and the negative gradient $-\nabla \distancefunction(\point^{t})$, which indicates the steepest
direction for distance $\distancefunction(\point^{t})$ decrease.
The process is then repeated $T$ times.

While this process can certainly refine $\interpolatedpointcloud$ to align with the ground truth high-res point cloud $\groundtruthpointcloud$. In practice, obviously $\groundtruthpointcloud$ is not available during inference, which means it is not possible to compute $\distancefunction$. Therefore a differentiable approximation of $\distancefunction$ is required.

\begin{figure*}[t!]
\centering
\includegraphics[width=0.9\textwidth]{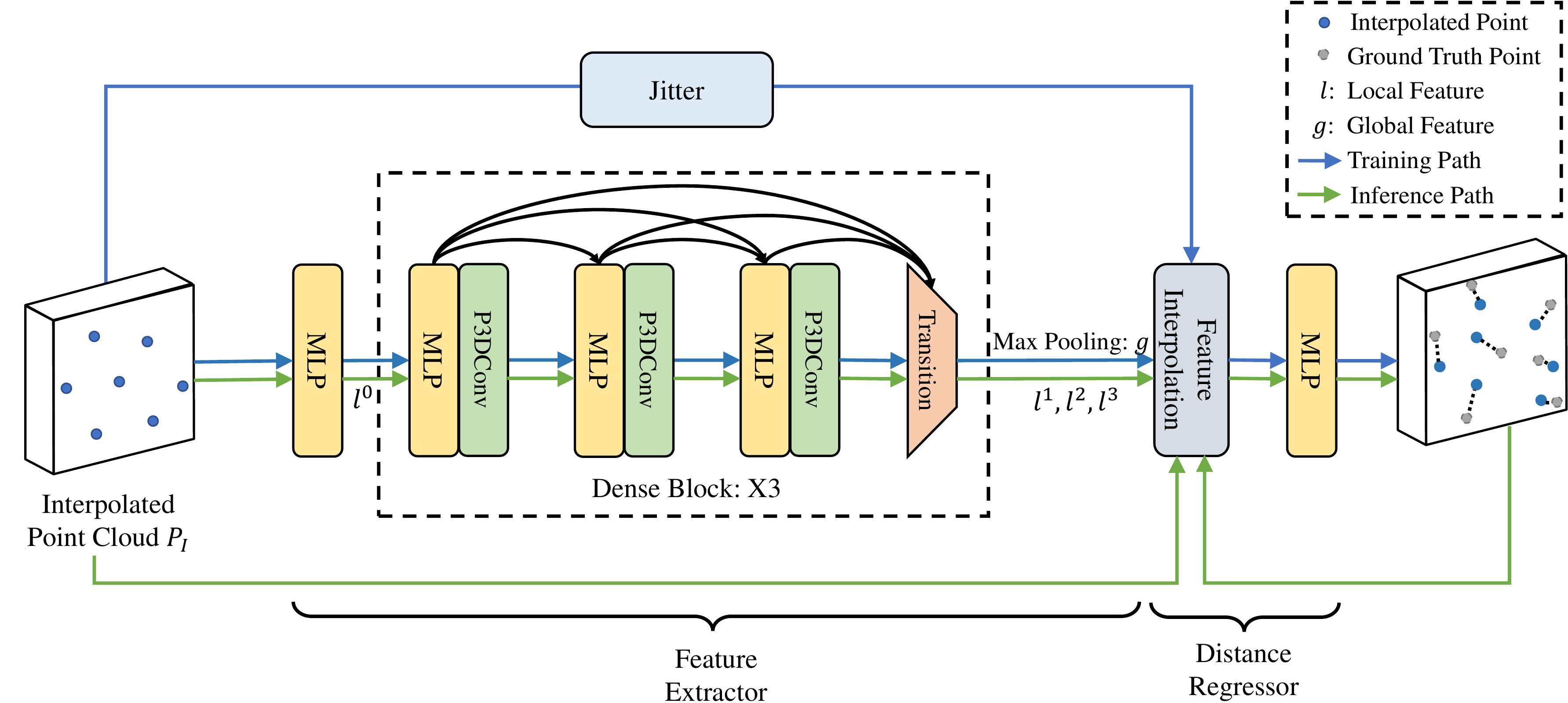}
\vspace{-0.1in}
\caption{Our P2PNet contains two submodules: a feature extractor and point-to-point distance regressor. For feature extractor, we stack an initial MLP and three dense blocks with intra-block dense connection \cite{huang2017densely}, where each dense block has three convolution groups to capture local features and one transition down layer to reduce channel. In distance regressor, we estimate the point-to-point distance for each query point conditioned on the extracted local and global features. Best viewed in color.\label{fig:p2pnet}}
\end{figure*}

\subsection{P2PNet for Distance Function Learning}
\label{sec:p2pnet}
We design a Point-to-Point Distance Network (P2PNet) to approximate $\distancefunction$ and serve as the oracle. 
In this section, we detail the design of P2PNet which mainly consists of a feature extractor and distance regressor, as in \Fig{p2pnet}.

\paragraph{Feature Extractor}
To capture the local and global geometric information of irregular points, we adopt the Point 4D Convolution from P4Transformer~\cite{fan2021point}, but simplify it to apply on the spatial domain only, thus termed as Point 3D Convolution (P3DConv).
Specifically, for each interpolated point $\point \in \interpolatedpointcloud$ and its associated feature $\feature$, we first search for its $k$-nearest neighbor $(\point_k, \feature_k)$, and calculate the coordinate offset $\displacement$ between them for convolution kernel generation.
Then the P3DConv on interpolated point $(\point, \feature)$ is conducted as follow:
\begin{equation}
    \feature^{\prime}=\sum_{\point_k \in K(\point)} \gamma(\alpha\displacement \odot \beta(\feature_k))
\end{equation}
where $\feature^{\prime}$ is the convoluted feature, $K(\point)$ is the $k$-nearest neighbor set of point $\point$, $\gamma, \alpha, \beta$ all indicate an MLP-based transformation with the same output channel $\dimension$, and $\odot$ represents the Hadamard product.

The detailed structure of our feature extractor is shown in \Fig{p2pnet}.
Given an interpolated point cloud $\interpolatedpointcloud \in \mathbb{R}^{N\times3}$, where $N$ is the number of points, an MLP first projects $\interpolatedpointcloud$ to a higher dimensional space $\mathbb{R}^{N\times \dimension}$, followed by a stack of three dense blocks with intra-block dense connection \cite{huang2017densely}. Each dense block consists of three convolution groups, followed by a transition down layer. Inside each convolutional group, an MLP reduces the feature dimension, while a P3DConv layer extracts local features.
The transition down layer is another MLP that reduces the feature channels and therefore following computational cost.
All MLPs for feature extraction share the same output channel of $\dimension$.
In the end, a set of multi-scale local features $\{\localfeature^0, \localfeature^1, \localfeature^2, \localfeature^3\} \in \mathbb{R}^{N\times \dimension}$ is captured.
By further applying a max pooling layer, a global feature $\globalfeature \in \mathbb{R}^{1 \times \dimension}$ is obtained.

\paragraph{Distance Regressor}
For any query point $\point \in \mathbb{R}^3$, its point-to-point distance $F(\point)$ is estimated based on the extracted local features $\{\localfeature^0, \localfeature^1, \localfeature^2, \localfeature^3\}$ and global feature $\globalfeature$.

To obtain the point-wise local features for each query point $\point$, 
we follow \cite{qi2017pointnet++} to 
conduct the feature interpolation, using the inverse distances of three-nearest neighbors in the initial interpolated point cloud as weights.
With that, the point-to-point distance $\distancefunction(\point)$ can be estimated as follow:
\begin{equation}
    \distancefunction(\point)\approx P2PNet(\point) = \psi (\point, \localfeature_{\point}^0, \localfeature_{\point}^1, \localfeature_{\point}^2, \localfeature_{\point}^3, \globalfeature)
    \label{eq:distanceregressor}
\end{equation}
where $\{\localfeature_{\point}^0, \localfeature_{\point}^1, \localfeature_{\point}^2, \localfeature_{\point}^3\}$ are the interpolated multi-scale local features, $\globalfeature$ is the global feature, and $\psi$ is a four-layer MLP.

\paragraph{Inference} During inference, the extracted local and global features are fixed. However in each iteration, since the points have moved, interpolated features are re-generated, thus new $\distancefunction(\point)$ can be estimated.

\paragraph{Training}
Unlike inference, there is no iterative optimization during training. Instead the interpolated points are jittered with Gaussian noise $\mathcal N(0, 0.02^2)$ to serve as query points, 
which simulates varying degrees of displacement in different iterations, and increases the smoothness and continuity of learned distance functions.

\paragraph{Loss Function}
We apply L1 loss to minimize the error between the predicted distance $P2PNet(\point)$ and the ground truth $F(\point)$.
\begin{equation}
    L(\interpolatedpointcloud) = \frac{1}{|\interpolatedpointcloud|} \sum_{\point \in \interpolatedpointcloud}|F(\point)-P2PNet(\point)|
\end{equation}

%% file: 5_eval.tex
\begin{figure*}[t!]
\centering
\includegraphics[width=\textwidth]{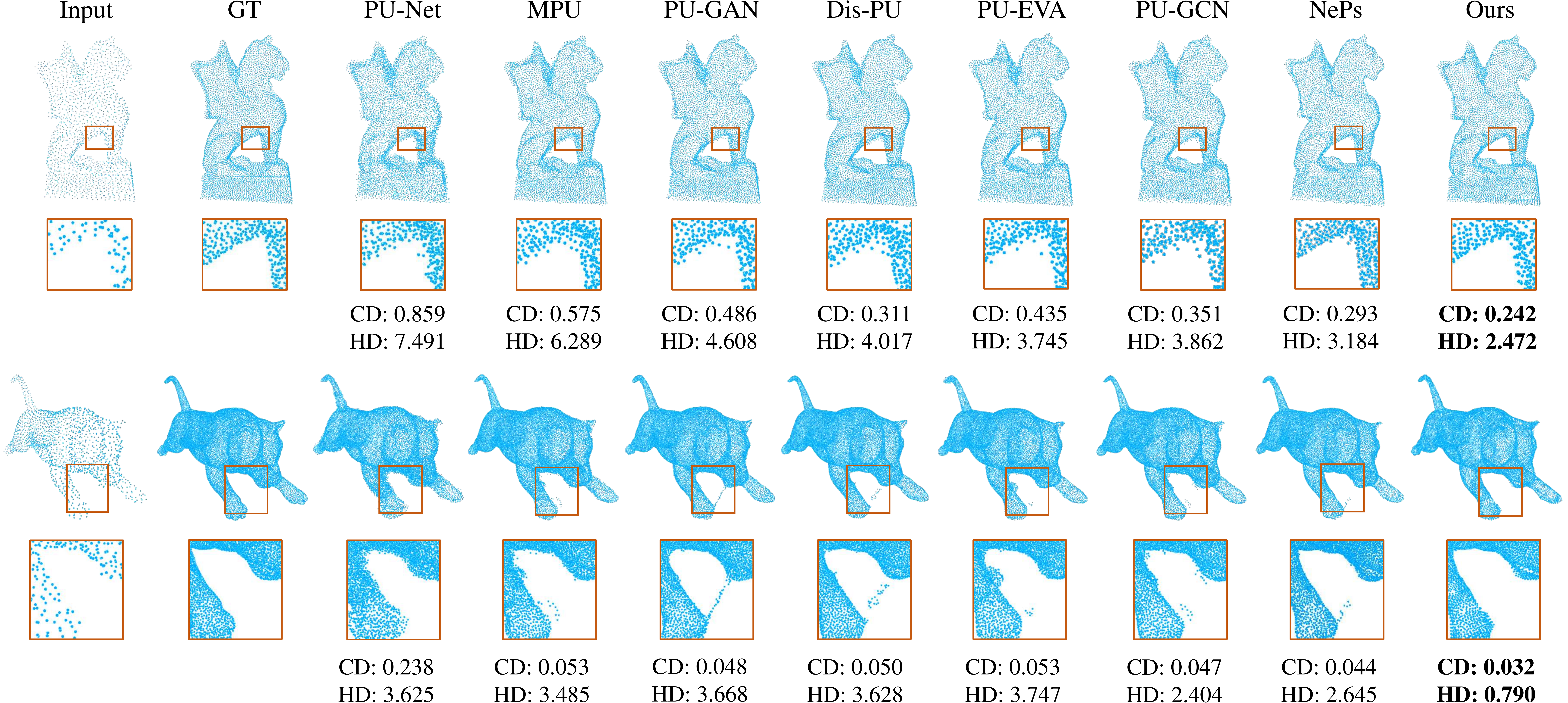}
\caption{Qualitative results on the PU-GAN dataset, where the first row is $4\times$ evaluation, the second row is $16\times$ evaluation. 
And the units of both CD and HD metrics are $10^{-3}$.
Our results clearly achieve the highest upsampling quality, with less outliers, smoother surfaces and more fine-grained details.}
\label{fig:pugan}
\vspace{-5pt}
\end{figure*}

\section{Evaluation}
In this section, we first demonstrate the superior performance of our algorithm against prior state-of-the-arts on public datasets.
And then validate the performance gain on downstream applications. Stress test results are also reported to demonstrate the robustness.
Finally, we provide comprehensive ablation studies to prove the effectiveness of each component.
Please refer to the supplementary material for 
implementation details and more comparative results.

\subsection{Experiment Setup}
\paragraph{Datasets}
Two public datasets, PU-GAN \cite{li2019pu} and PU1K \cite{qian2021pu} are used for evaluation. We follow the official training/testing splits and settings in original papers,
where training is conducted on patch level.
Compared to the PU-GAN dataset, PU1K is more challenging because it has a larger volume of data and more diverse categories.

During training, each input low-res patch 
contains $256$ points, while its corresponding high-res patch has $1024$ points. Thus the upsampling rate $\upsamplingrate =4$.
During testing, we follow~\cite{qian2021pu} to generate point clouds from the test set of PU-GAN with Poisson disk sampling \cite{yuksel2015sample}, as it only provides 3D meshes.
All testing low-res point clouds from both datasets have $2048$ points, while the high-res counterparts contain $2048 \times \upsamplingrate$ points.
Given the low-res input, we first generate the interpolated point cloud by our midpoint interpolation.
Then we follow \cite{qian2021pu} to extract patches, apply gradient descent to update them, and finally merge them to obtain the full high-res output.

Besides aboved synthetic datasets, we also adopt the real-scanned ScanObjectNN \cite{uy2019revisiting} and KITTI \cite{geiger2013vision} datasets for qualitative evaluation.

\paragraph{Evaluation Metrics}
Following \cite{li2021point,qian2021pu,luo2021pu}, we adopt the Chamfer distance (CD), Hausdorff distance (HD) and point-to-surface distance (P2F) as metrics.

\paragraph{Baselines}
For the PU-GAN dataset, we train PU-Net \cite{yu2018pu}, MPU \cite{yifan2019patch}, PU-GAN \cite{li2019pu}, Dis-PU \cite{li2021point}, PU-EVA \cite{luo2021pu}, PU-GCN \cite{qian2021pu} and Neural Points (NePs) \cite{feng2022neural} with the default settings in the respective papers as baselines.
For the PU1K dataset, 
we only choose PU-Net \cite{yu2018pu}, MPU \cite{yifan2019patch} PU-GCN \cite{qian2021pu} and PU-Transformer \cite{qian2021pu}, following original papers.

\begin{figure*}[t!]
\vspace{-5pt}
\centering
\includegraphics[width=\textwidth]{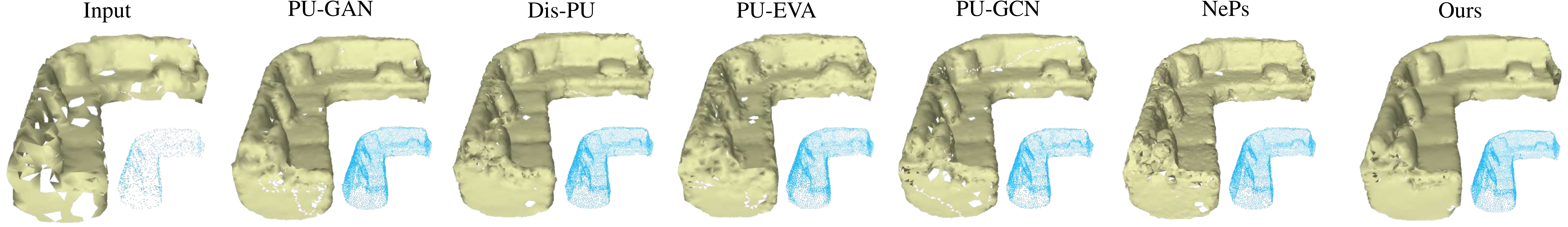}
\vspace{-10pt}
\caption{$4\times$ upsampled results on the ScanObjectNN dataset, and the meshes are reconstructed by BallPivoting surface reconstruction algorithm \cite{bernardini1999ball}.
Our method generates more complete, smooth and faithful mesh and point cloud.}
\label{fig:scanobjectnn}
\end{figure*}

\begin{figure*}[htbp]
\centering
\includegraphics[width=\textwidth]{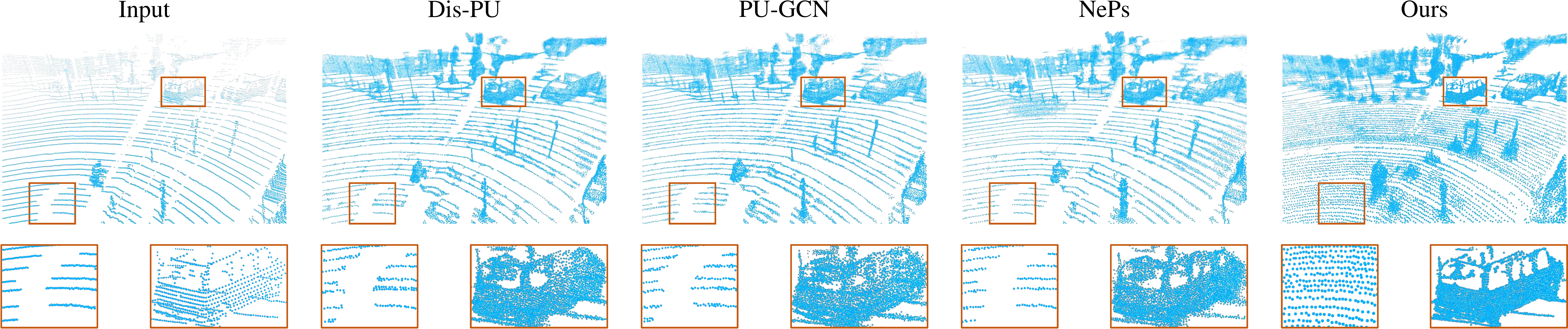}
\vspace{-10pt}
\caption{$4\times$ upsampled results on the KITTI dataset. Our result not only retains more fine-grained details but also fills the gaps between LiDAR fibers.}
\label{fig:kitti}
\vspace{-5pt}
\end{figure*}

\subsection{Comparison with SOTA}
\paragraph{Results on the PU-GAN Dataset}
\Tab{pugan} shows that our method outperforms prior arts on all metrics.
In particular, the performance gain on higher upsampling rate ($16\times$) is larger.
As shown in \Fig{pugan}, previous methods tend to generate outliers caused by overestimation of 3D coordinates. This artifact is more severe as the upsampling rate goes higher.
On the contrary, our results have much less outliers, more faithful surfaces and more fine-grained details, without obvious distinction between $4\times$ and $16\times$ upsampling.

\begin{table}[htbp]
\centering
\vspace{-5pt}
\resizebox{\linewidth}{!}{
\begin{tabular}{@{}lcccccccc@{}}
\toprule 
 Rates & \multicolumn{5}{@{}c@{}}{$4\times$ $(\upsamplingrate=4)$} & \multicolumn{3}{@{}c@{}}{$16\times$ $(\upsamplingrate=16)$} \\
  \cmidrule(lr){2-6} \cmidrule(lr){7-9}
Methods  & \begin{tabular}[c]{@{}c@{}} CD $\downarrow$ \\ $10^{-3}$ \end{tabular} & \begin{tabular}[c]{@{}c@{}} HD $\downarrow$ \\ $10^{-3}$ \end{tabular} & \begin{tabular}[c]{@{}c@{}} P2F $\downarrow$ \\ $10^{-3}$ \end{tabular} & \begin{tabular}[c]{@{}c@{}} Param. $\downarrow$  \\ Kb \end{tabular} & \begin{tabular}[c]{@{}c@{}} Time $\downarrow$ \\ s \end{tabular} & \begin{tabular}[c]{@{}c@{}} CD $\downarrow$ \\ $10^{-3}$ \end{tabular}  & \begin{tabular}[c]{@{}c@{}} HD $\downarrow$ \\ $10^{-3}$ \end{tabular}  & \begin{tabular}[c]{@{}c@{}} P2F $\downarrow$ \\ $10^{-3}$ \end{tabular} \\
\midrule
  PU-Net \cite{yu2018pu}  & 0.529 & 6.805 & 4.760 & 814.3 & 0.566 & 0.510 
  & 8.206  & 6.041 \\
  MPU  \cite{yifan2019patch} & 0.292 & 6.672 & 2.822 & 76.2 & 0.573 & 0.219  & 7.054  & 3.085 \\
  PU-GAN \cite{li2019pu} & 0.282 & 5.577 & 2.016 & 684.2 & 0.698 & 0.207  
  & 6.963  & 2.556 \\
  Dis-PU \cite{li2021point}  & 0.274 & 3.696 & 1.943 & 1047.0 & 1.604 & 0.167  & 4.923  & 2.261 \\
  PU-EVA \cite{luo2021pu}   & 0.277 & 3.971 & 2.524 & 2869.0 & 0.740 & 0.185  & 5.273  & 2.972 \\
  PU-GCN \cite{qian2021pu}  & 0.268 & 3.201 & 2.489 & 76.0 & 0.538 & 0.161  & 4.283  & 2.632 \\
  NePs \cite{feng2022neural} & 0.259 & 3.648 & 1.935 & 664.1 & 0.403 & 0.152 & 4.910 & 2.198 \\
\hline
  \textbf{Ours} &\textbf{0.245} & \textbf{2.369} 
  & \textbf{1.893} & \textbf{67.1} & \textbf{0.384} & \textbf{0.108}  & \textbf{2.352}  & \textbf{2.127}\\
\bottomrule
\end{tabular}}
\vspace{-5pt}
\caption{Quantitative results on the PU-GAN dataset, where $4\times$ and $16\times$ represent the upsampling rate $\upsamplingrate=4$ and $\upsamplingrate=16$ respectively.
Our method outperforms others in both accuracy and efficiency. 
}
\label{tab:pugan}
\vspace{-5pt}
\end{table}

In addition, we compare the efficiency of each method under $4\times$ setting, in terms of network parameters (Param.), as well as inference time which is measured end-to-end from loading the input low-res point cloud to generate the full high-res output, using a TITAN X GPU. Note that model size and inference time are not necessarily in proportion, because some of the expensive operations, \eg $k$-nearest neighbor search, are not part of the network. That said, our method is the fastest with the fewest parameters.

\paragraph{Arbitrary Upsampling Rates}
Unlike most of previous methods \cite{yu2018pu,li2019pu,yifan2019patch,li2021point,qian2021pu,qian2020pugeo,long2022pc2,qiu2021pu}, our approach does not need retraining for different upsampling rates.
Similarly, priort art NePs \cite{feng2022neural} also does not require retraining. Thus we conduct a comparison using the model trained on PU-GAN dataset.
For both methods, we only vary the upsampling rate $\upsamplingrate$ during inference while fixing all other parameters.
Based on \Tab{arbitraryscale}, our method yields better accuracy under most of the metrics, except the P2F metric with $6\times$ and $7\times$ upsampling.
Note that since P2F is asymmetrical, only from upsampled points to the ground truth surfaces but not vice versa\cite{li2019pu}, CD and HD metrics are more meaningful.

\begin{table}[htbp]
\centering
\vspace{-5pt}
\resizebox{.8\linewidth}{!}{
\small
\begin{tabular}{@{}lcccccc@{}}
\toprule 
Methods & \multicolumn{3}{@{}c@{}}{NePs \cite{feng2022neural}} & \multicolumn{3}{@{}c@{}}{Ours} \\
  \cmidrule(lr){2-4} \cmidrule(lr){5-7}
  
Rates  & \begin{tabular}[c]{@{}c@{}} CD $\downarrow$ \\ $10^{-3}$ \end{tabular} & \begin{tabular}[c]{@{}c@{}} HD $\downarrow$ \\ $10^{-3}$ \end{tabular} & \begin{tabular}[c]{@{}c@{}} P2F $\downarrow$ \\ $10^{-3}$ \end{tabular} & \begin{tabular}[c]{@{}c@{}} CD $\downarrow$ \\ $10^{-3}$ \end{tabular}  & \begin{tabular}[c]{@{}c@{}} HD $\downarrow$ \\ $10^{-3}$ \end{tabular}  & \begin{tabular}[c]{@{}c@{}} P2F $\downarrow$ \\ $10^{-3}$ \end{tabular} \\
  
\midrule
  $2\times$  & 0.642 & 7.324 & 2.574 & \textbf{0.540} & \textbf{3.177}  & \textbf{1.775} \\
  $3\times$ & 0.409 & 5.389 & 2.176 & \textbf{0.353}  & \textbf{2.608}  & \textbf{1.654} \\
  $5\times$ & 0.248 & 3.922 & 1.871 & \textbf{0.234}  & \textbf{2.549}  & \textbf{1.836} \\
  $6\times$  & 0.242 & 3.671 & \textbf{1.809} & \textbf{0.225}  & \textbf{2.526}  & 1.981 \\
  $7\times$   & 0.237 & 3.796 & \textbf{1.795} & \textbf{0.219}  & \textbf{2.634}  & 1.940 \\
\bottomrule
\end{tabular}}
\vspace{-5pt}
\caption{NePs \cite{feng2022neural} \vs ours on the PU-GAN dataset with upsampling rate $\upsamplingrate \in \{2,3,5,6,7\}$. 
Our method achieves superior accuracy across most cases.
}
\label{tab:arbitraryscale}
\vspace{-5pt}
\end{table}

\paragraph{Results on the PU1K Dataset}
We also conduct the $4\times$ evaluation on the more challenging
PU1K dataset, as reported in \Tab{pu1k}.
Our method still outperforms others on almost all metrics, except for the P2F metric, which is second to PU-Transformer \cite{qiu2021pu}.
Note that our model is much smaller than PU-Transformer (67.1Kb \vs 969.9Kb).

\begin{table}[htbp]
\centering
\resizebox{.7\linewidth}{!}{
\begin{tabular}{lccc}
\toprule    
Methods & \begin{tabular}[c]{@{}c@{}} CD $\downarrow$ \\ $10^{-3}$ \end{tabular} & \begin{tabular}[c]{@{}c@{}} HD $\downarrow$ \\ $10^{-3}$ \end{tabular} & \begin{tabular}[c]{@{}c@{}} P2F $\downarrow$ \\ $10^{-3}$ \end{tabular} \\
\midrule
PU-Net \cite{yu2018pu} &  1.155 & 15.170 & 4.834 \\
MPU \cite{yifan2019patch} & 0.935 & 13.327 & 3.511 \\
PU-GCN \cite{qian2021pu} & 0.585 & 7.577 & 2.499 \\
PU-Transformer \cite{qiu2021pu} & 0.451 & 3.843 & \textbf{1.277} \\
\hline
  \textbf{Ours} &\textbf{0.404} & \textbf{3.732} 
  & 1.474 \\
\bottomrule
\end{tabular}}
\vspace{-5pt}
\caption{\label{tab:pu1k} $4\times$ quantitative results on the PU1K dataset, where the results of other methods are directly borrowed from the original papers. Our method outperforms others on nearly all metrics.}
\vspace{-5pt}
\end{table}

\paragraph{Results on Real Datasets}
Using models trained on the PU-GAN dataset, we also conduct experiments on real-scanned point clouds from ScanObjectNN \cite{uy2019revisiting} and KITTI \cite{geiger2013vision} datasets, as shown in \Fig{scanobjectnn} and \Fig{kitti}.
Since there is no ground truth high-res point cloud, we only qualitatively compare, and omit some methods that produce consistently worse results.
Not only sparse and noisy, scanned data often have small holes or gaps, which makes them even more challenging.
\Fig{scanobjectnn} shows that our results are more complete, smooth and faithful, while other methods tend to keep the holes.
In \Fig{kitti}, our result appears to be more complete and with more fine-grained details.

\subsection{Impact on Downstream Tasks}
We further highlight the upsampling quality in two downstream applications: point cloud classification and surface reconstruction.

\paragraph{Point Cloud Classification}
We adopt CurveNet~\cite{xiang2021walk} as the classification model, and utilize the same training and testing schema on the ModelNet40 dataset~\cite{wu20153d}.
Specifically, the model is trained with 1024 points. 
For each testing point cloud, we uniformly subsample 256 points as the low-res input, and upsample them back to 1024 points with various methods (trained on the PU-GAN dataset).

We then compare the classification performance on the downsampled low-res point clouds (Low-res, 256 points), original test set (High-res, 1024 points) and upsampled point clouds (1024 points).
As shown in the first two rows of \Tab{downstream}, the classification accuracy of the low-res point clouds is observably worse, while our upsampling method brings a significant improvement.

\begin{table}[htbp]
\centering
\vspace{-5pt}
\resizebox{\linewidth}{!}{
\begin{tabular}{@{}clllll@{}}
\toprule
\multirow{4}{*}{\begin{tabular}[c]{@{}c@{}}Classification \\ Accuracy (\%) $\uparrow$ \end{tabular}} & Low-res & High-res & PU-Net \cite{yu2018pu}  & MPU \cite{yifan2019patch}  & PU-GAN \cite{li2019pu} \\
& 68.76   & 93.72    & 88.82  & 89.91  & 90.25   \\
\cmidrule(l){2-6} 
& Dis-PU \cite{li2021point}  & PU-EVA \cite{luo2021pu}  & PU-GCN \cite{qian2021pu}  & NePs \cite{feng2022neural}  & Ours \\
& 91.57   & 90.83    & 91.21  & 91.39   &\textbf{91.96} \\ \midrule
\multirow{4}{*}{\begin{tabular}[c]{@{}c@{}} Reconstruction \\CD  ($10^{-3}$) $\downarrow$ \end{tabular}} & Low-res & High-res & PU-Net \cite{yu2018pu}  & MPU \cite{yifan2019patch}  & PU-GAN \cite{li2019pu} \\
& 0.106   & 0.039    & 0.221  & 0.102  & 0.090  \\
\cmidrule(l){2-6} 
& Dis-PU \cite{li2021point}  & PU-EVA \cite{luo2021pu}  & PU-GCN \cite{qian2021pu}  & NePs \cite{feng2022neural}  & Ours \\
& 0.084   & 0.086    & 0.079  & 0.075  & \textbf{0.071} \\ \bottomrule
\end{tabular}}
\vspace{-5pt}
\caption{Results on downstream tasks. The first two rows are the overall classification accuracy on ModelNet40, and the last two rows measure the surface reconstruction error with Chamfer distance on the PU-GAN dataset. 
``Low-res'' denotes the downsampled point clouds, 
and ``High-res'' denotes the high-res counterparts.
Obviously our upsampled point clouds bring the most significant performance improvement to downstream tasks.}
\label{tab:downstream} 
\vspace{-5pt}
\end{table}

\paragraph{Surface Reconstruction}
We utilize BallPivoting \cite{bernardini1999ball} to reconstruct meshes from the $4\times$ upsampled point clouds (8192 points) of PU-GAN dataset. 
From the last two rows of \Tab{downstream}, we find that the low-res point clouds (Low-res, 2048 points) already achieve a comparable performance, because they are directly sampled from the ground truth meshes.
Although the improvement obtained by each method is marginal, our approach still yields the best.

\subsection{Robustness Test}
\begin{figure}[htbp]
\centering
\includegraphics[width=\linewidth]{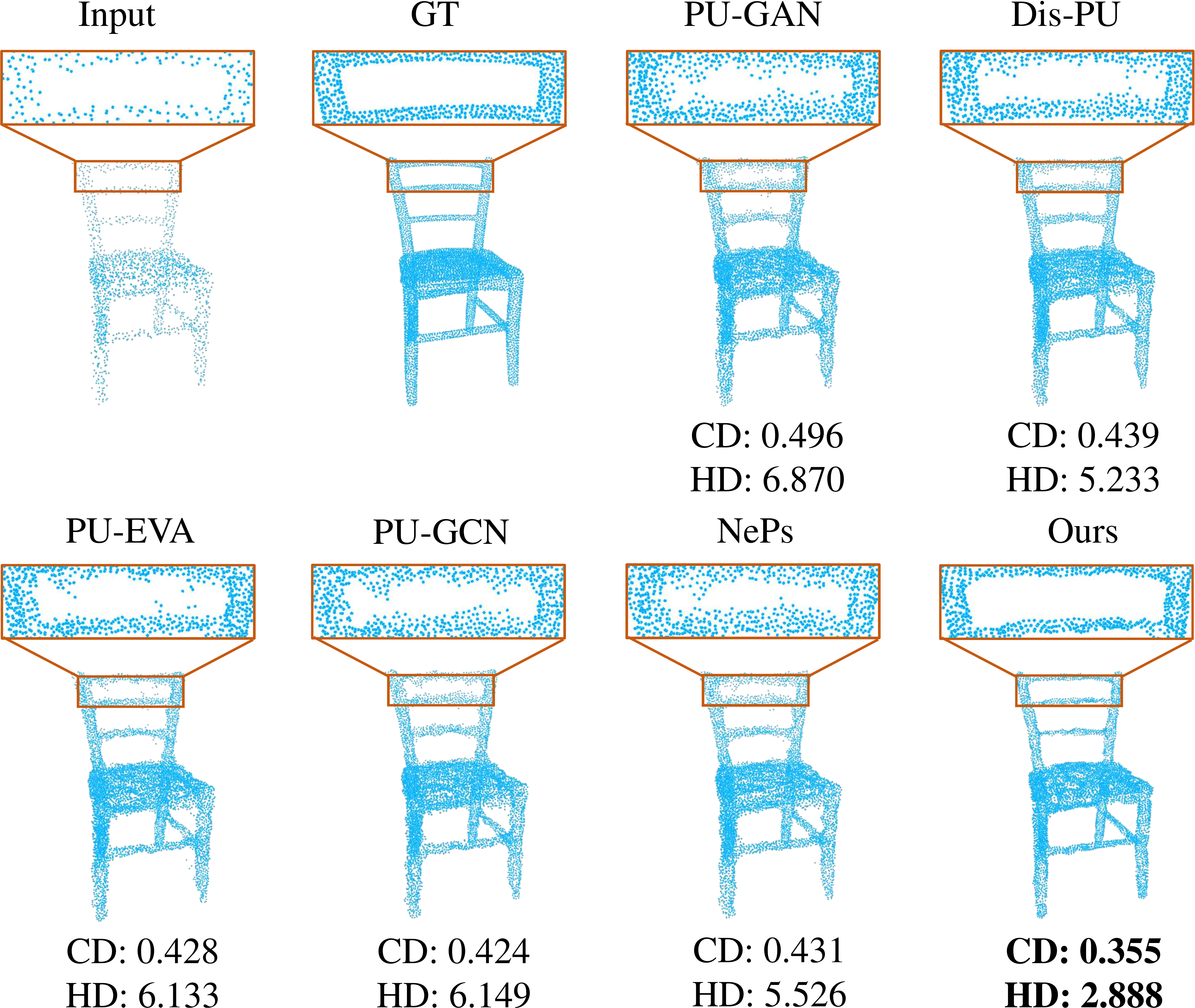}
\vspace{-10pt}
\caption{$4\times$ qualitative results on the PU-GAN dataset with added noise level $\noiselevel=0.01$. 
The units of both CD and HD metrics are $10^{-3}$.
Our result is cleaner with less outliers and higher fidelity.}
\vspace{-5pt}
\label{fig:noise}
\end{figure}

\paragraph{Additive Noise}
As the point clouds captured by scanners are often noisy, it is necessary to evaluate the robustness of each method against noise.
To be specific, we first generate some random noise offline, which is sampled from a standard Gaussian distribution $\mathcal{N}(0,1)$ and multiplied by a factor $\noiselevel$, where $\noiselevel$ denotes the noise level.
Then we test on the low-res point clouds of PU-GAN dataset with added noise.
And the training of all methods incorporate with Gaussian noise perturbation as augmentation strategy for fair comparison.
As shown in \Tab{noise}, our method achieves the best performance consistently, especially at high noise level.
And \Fig{noise} provides the qualitative comparisons, which verifies that our result is cleaner with much less outliers.

\begin{table}[htbp]
\centering
\resizebox{0.9\linewidth}{!}{
\small
\begin{tabular}{@{}lcccccc@{}}
\toprule 
Noise Levels & \multicolumn{3}{@{}c@{}}{$\noiselevel=0.01$} & \multicolumn{3}{@{}c@{}}{$\noiselevel=0.02$} \\
  \cmidrule(lr){2-4} \cmidrule(lr){5-7}
Methods   & \begin{tabular}[c]{@{}c@{}} CD $\downarrow$ \\ $10^{-3}$ \end{tabular} & \begin{tabular}[c]{@{}c@{}} HD $\downarrow$ \\ $10^{-3}$ \end{tabular} & \begin{tabular}[c]{@{}c@{}} P2F $\downarrow$ \\ $10^{-3}$ \end{tabular}  & \begin{tabular}[c]{@{}c@{}} CD $\downarrow$ \\ $10^{-3}$ \end{tabular} & \begin{tabular}[c]{@{}c@{}} HD $\downarrow$ \\ $10^{-3}$ \end{tabular} & \begin{tabular}[c]{@{}c@{}} P2F $\downarrow$ \\ $10^{-3}$ \end{tabular} \\
\midrule
  PU-Net \cite{yu2018pu}  & 0.628 & 8.068 & 9.816 & 1.078 
  & 10.867  & 16.401 \\
  MPU  \cite{yifan2019patch} & 0.506 & 6.978 & 9.059 & 0.929  & 10.820  & 15.621 \\
  PU-GAN \cite{li2019pu} & 0.464 & 6.070 & 7.498 & 0.887 
  & 10.602  & 15.088 \\
  Dis-PU \cite{li2021point}  & 0.419 & 5.413 & 6.723 & 0.818  & 9.345  & 14.376 \\
  PU-EVA \cite{luo2021pu}   & 0.459 & 5.377 & 7.189 & 0.839  & 9.325  & 14.652 \\
  PU-GCN \cite{qian2021pu}  & 0.448 & 5.586 & 6.989 & 0.816  & 8.604  & 13.798 \\
  NePs \cite{feng2022neural} & 0.425 & 5.438 & 6.546 & 0.798 & 9.102 & 12.088 \\
\hline
  \textbf{Ours} &\textbf{0.414} & \textbf{4.145} 
  & \textbf{6.400} & \textbf{0.766}  & \textbf{7.339}  
  & \textbf{11.534}\\
\bottomrule
\end{tabular}}
\vspace{-5pt}
\caption{$4\times$ quantitative results on the PU-GAN dataset with different noise level $\noiselevel$. It is obvious that our method consistently surpasses all other approaches. }
\label{tab:noise}
\vspace{-5pt}
\end{table}

\begin{figure}[htbp]
\centering
\includegraphics[width=\linewidth]{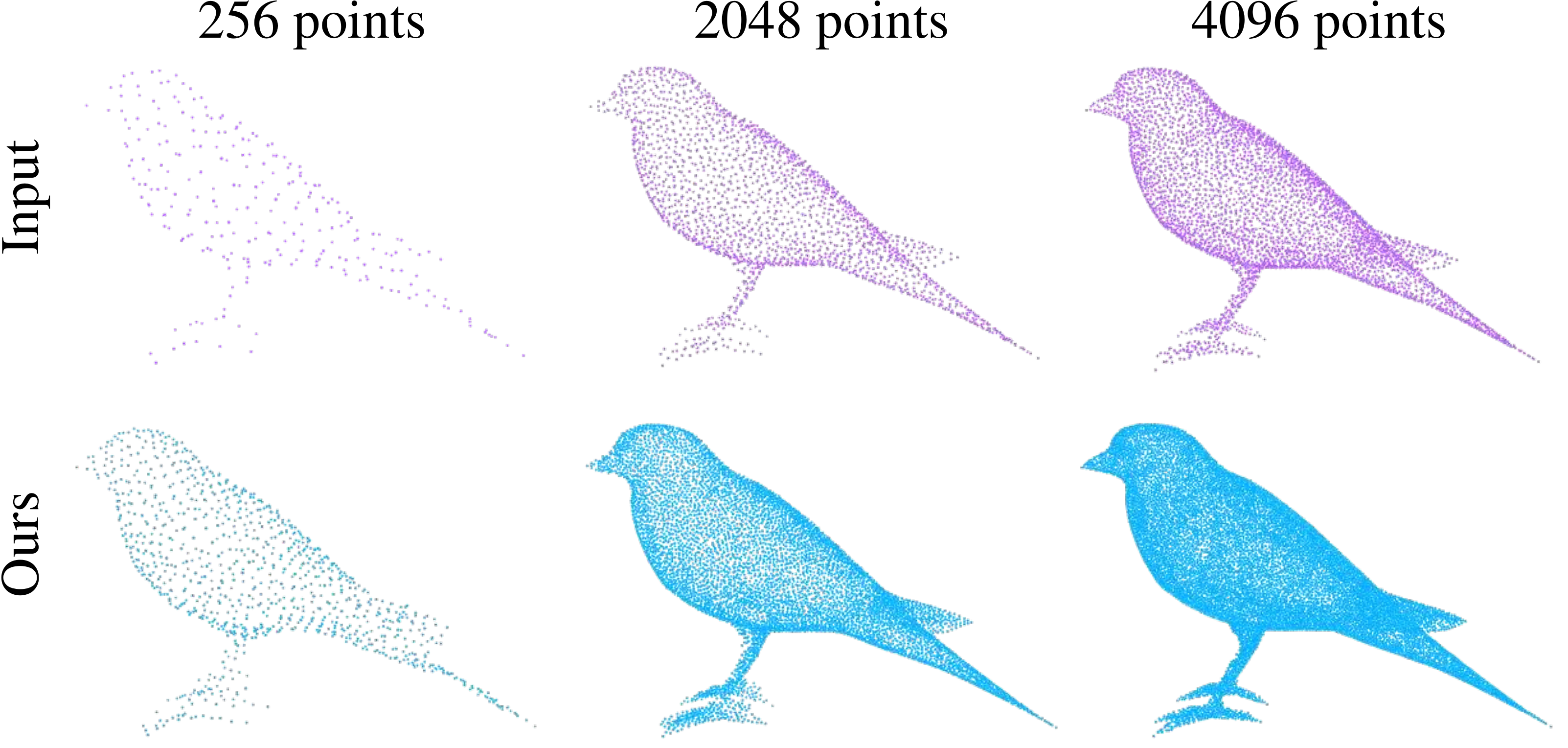}
\vspace{-0.1in}
\caption{$4\times$ upsampled results by our method with different input sizes. Our method consistently achieves high upsampling quality regardless of the input density.}
\label{fig:inputsize}
\vspace{-5pt}
\end{figure}

\paragraph{Various Input Sizes}
Considering all previous evaluations of PU-GAN and PU1K datasets are conducted on the low-res point clouds with 2048 points, we further validate the robustness of our method against different input sizes.
As \Fig{inputsize} shows, although our model is trained on the fixed-scale input data, it can generalize well to different scales during inference, even when the input is extremely sparse.

\subsection{Ablation Study}
In this section, we conduct the comprehensive ablation studies to validate the effectiveness of each component, and all these experiments are based on the PU1K dataset with $4\times$ setting.

\paragraph{Distance Regression \vs Coordinate Prediction}
Compared with predicting 3D coordinates or residuals, merely regressing the point-to-point distance is a relatively easier task.
To verify this point, we modify the output channel of P2PNet's last MLP to predict the 3D coordinate offset for each interpolated point, and we employ a L2 loss to minimize the prediction error.
Following \cite{luo2021score,xiang2017posecnn,ma2020deep}, we update the interpolated points to ground truth locations in an auto-regression way:
\begin{equation}
\begin{split}
    & \point^{t+1}=\point^{t}+\stepsize \Delta (\point^{t}),
    t=0,...,\iteration-1
    \label{eq:autoregressive}
\end{split}
\end{equation}
where $\Delta(\point^{t})$ denotes the predicted coordinate displacement, $\stepsize$ is the step size, and we fine-tune it and iteration number $\iteration$ to achieve the best performance, holding all other parameters the same.
Moreover, we also report the results of end-to-end update, which repeats \Eq{autoregressive} only once with step size $\stepsize=1$.

\begin{table}[htbp]
\centering
\resizebox{\linewidth}{!}{
\begin{tabular}{lccc}
\toprule    
Prediction Contents & \begin{tabular}[c]{@{}c@{}} CD $\downarrow$ \\ $10^{-3}$ \end{tabular} & \begin{tabular}[c]{@{}c@{}} HD $\downarrow$ \\ $10^{-3}$ \end{tabular} & \begin{tabular}[c]{@{}c@{}} P2F $\downarrow$ \\ $10^{-3}$ \end{tabular} \\
\midrule
3D Coordinate Offset (End-to-end) &  1.170 & 11.834 & 2.521 \\
3D Coordinate Offset (Auto-regression) & 0.663 & 7.034 & 1.935 \\
\hline
  \textbf{Point-to-point Distance} &\textbf{0.404} & \textbf{3.732} 
  & \textbf{1.474} \\
\bottomrule
\end{tabular}}
\caption{\label{tab:prediction} $4\times$ comparative results on the PU1K dataset with different prediction contents. Predicting distance obviously achieves the best performance.}
\end{table}

As reported in \Tab{prediction}, our distance regression based method clearly achieves superior performance, although updating points auto-regressively can alleviate the misestimation of predicted coordinate offsets to some extent.

\paragraph{Midpoint Interpolation}
We employ the midpoint interpolation result $\interpolatedpointcloud$ to serve as both input of P2PNet and initial point cloud to be updated.
For validating the effectiveness of midpoint interpolation, we replace the input of P2PNet with the low-res point cloud $\inputpointcloud$. 
And we also sample points from the Gaussian distribution $\mathcal{N}(\point, \sigma^2)$ to get another initial point cloud, denoted as $P_S$.
Finally, we combine different network inputs and initial point clouds to conduct the experiments, holding all other parameters the same.

\begin{table}[htbp]
\centering
\resizebox{\linewidth}{!}{
\begin{tabular}{llccc}
\toprule    
 Network Inputs &  Initial Point Clouds to Be Updated   & \begin{tabular}[c]{@{}c@{}} CD $\downarrow$ \\ $10^{-3}$ \end{tabular} & \begin{tabular}[c]{@{}c@{}} HD $\downarrow$ \\ $10^{-3}$ \end{tabular} & \begin{tabular}[c]{@{}c@{}} P2F $\downarrow$ \\ $10^{-3}$ \end{tabular} \\
\midrule
$\inputpointcloud$ & $P_S$ &  0.713 & 5.586 & 1.656 \\
$\inputpointcloud$ & $\interpolatedpointcloud$ & 0.426 & 3.989 & 1.813 \\
$\interpolatedpointcloud$ & $P_S$ &  0.654 & 5.295 & 1.616 \\
\hline
  \bm{$\interpolatedpointcloud$} & \bm{$\interpolatedpointcloud$} &\textbf{0.404} & \textbf{3.732} & \textbf{1.474} \\
\bottomrule
\end{tabular}}
\caption{\label{tab:interpolation} $4\times$ comparative results on the PU1K dataset with different combinations of network input and initial point cloud. Using $\interpolatedpointcloud$ as both network input and initial point cloud clearly achieves the best perfomance.}
\vspace{-5pt}
\end{table}

From the second and forth rows of \Tab{interpolation}, we conclude that using the denser interpolated point cloud $\interpolatedpointcloud$ as network input can achieve better performance, since it contains more geometrical information and benefits the feature extraction process.
And the improvement is more evident in the comparison between the third and forth rows, it proves that our midpoint interpolation result provides a better initial position, which benefits the update process under the same number of iterations.

\paragraph{P3DConv \vs EdgeConv \cite{yifan2019patch}}
The EdgeConv based feature extractor \cite{yifan2019patch} is widely used by previous work \cite{li2019pu,li2021point,yifan2019patch,feng2022neural,zhao2022self}.
However, EdgeConv \cite{yifan2019patch} utilizes most of parameters to refine each individual feature.
While our P3DConv focuses on the feature aggregation achieved by generated convolution kernels, thus benefits the extraction of local and global features.
For verifying the superior performance of our P3DConv, we replace the dense block in our P2PNet with EdgeConv in \cite{yifan2019patch}, and we also fine-tune the number of EdgeConv layers to achieve the comparable network parameters, as \Tab{point3dconv} shows.

\begin{table}[htbp]
\centering
\resizebox{.88\linewidth}{!}{
\begin{tabular}{lcccc}
\toprule    
Convolution Layers & \begin{tabular}[c]{@{}c@{}} CD $\downarrow$ \\ $10^{-3}$ \end{tabular} & \begin{tabular}[c]{@{}c@{}} HD $\downarrow$ \\ $10^{-3}$ \end{tabular} & \begin{tabular}[c]{@{}c@{}} P2F $\downarrow$ \\ $10^{-3}$ \end{tabular} & \begin{tabular}[c]{@{}c@{}} Param. $\downarrow$ \\ Kb \end{tabular} \\
\midrule
EdgeConv \cite{yifan2019patch} &  0.878 & 17.645 & 3.363 & 71.2 \\
  \textbf{P3DConv} &\textbf{0.404} & \textbf{3.732} 
  & \textbf{1.474} & \textbf{67.1} \\
\bottomrule
\end{tabular}}
\caption{\label{tab:point3dconv} $4\times$ comparative results on the PU1K dataset with different convolution layers. Our P3DConv is clearly more effective than EdgeConv \cite{yifan2019patch}.}
\vspace{-5pt}
\end{table}

%% file: 6_conclusion.tex
\section{Conclusion}
We propose a novel method for precise point cloud upsampling, supporting arbitrary upsampling rates after training once.
For arbitrary upsampling rates, we propose to directly upsample points in Euclidean space via midpoint interpolation and then refine them, which decouples the point generation from network learning.
For refining the interpolated points more precisely, we regard the refinement as an optimization problem, and then solve it by minimizing the learned point-to-point distance function.
And considering the ground truth point cloud is not avaliable during inference, we construct P2PNet to approximate the point-to-point distance function in a differentiable way.
Extensive quantitative and qualitative comparisons on benchmarks and downstream tasks demonstrate that our method outperforms prior state-of-the-art methods, while achieving the fewest parameters and fastest inference speed.

\paragraph{Acknowledgments}
This work was supported in part by NSFC Project (62176061) and STCSM Project (No.22511105000). 
Danhang Tang, Yinda Zhang and Xiangyang Xue are the corresponding authours.

%% file: 7_supp.tex
\noindent \textbf{\Large Supplementary Material}
\setcounter{section}{0}
\vspace{+10pt}

In this supplementary material, we provide additional implementation details, ablation studies and qualitative results.

\section{Implementation Details}
Additional details about hyperparameter settings and detailed network architecture are elucidated in this section.

\subsection{Hyperparameters}
In the P2PNet, we set the feature dimension $\dimension=32$ and the nearest neighbor number $k=16$.
For training, we use random perturbation, rotation and scaling for data augmentation, the same as \cite{qian2021pu}.
For testing, we choose the iteration number $\iteration=10$ to balance the computational cost and performance.

Our model is implemented with Pytorch \cite{paszke2019pytorch}, trained on a NVIDIA TITAN X GPU for 60 epochs with a batch size of 32.
We use the Adam optimizer \cite{kingma2014adam} with an initial learning rate of 1e-3 and a decay factor of 0.5 every 20 epochs.

\subsection{Detailed Network Architecture}
\begin{figure*}[t!]
\centering
\includegraphics[width=0.97\textwidth]{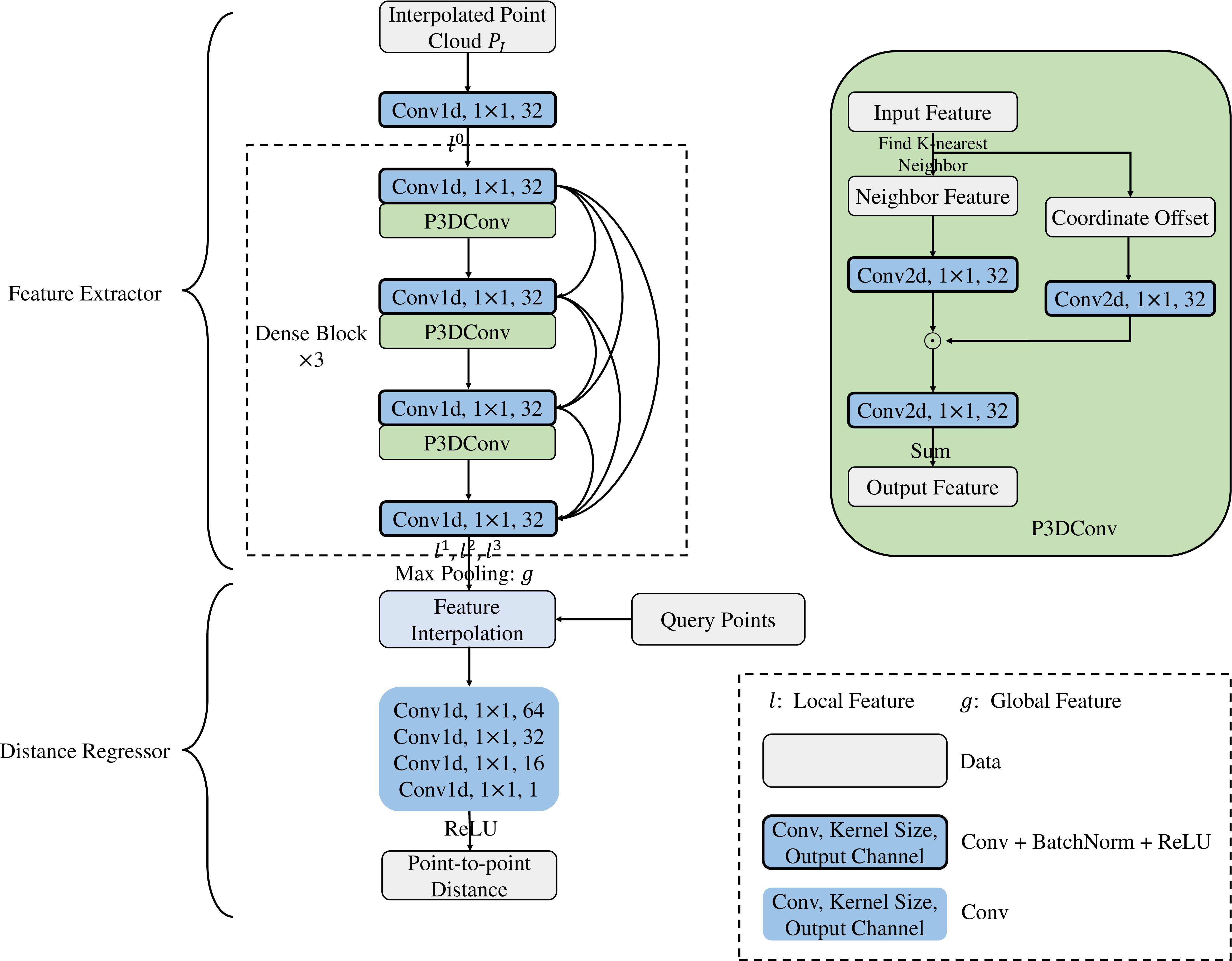}
\caption{The detailed structure of our P2PNet.}
\label{fig:structure}
\end{figure*}

The detailed architecture of our P2PNet is shown in \Fig{structure}.
In the feature extractor, we first employ an MLP to project the initial interpolated point cloud $\interpolatedpointcloud$ to a higher dimension space.
And then stack three dense blocks with intra-block dense connection \cite{huang2017densely}, where each block comprises four MLPs and three P3DConv layers.
The MLP is used to reduce feature channel, and the P3DConv layer is utilized for local feature capture.
Lastly, we obtain the extracted local features $\{\localfeature^0,\localfeature^1,\localfeature^2,\localfeature^3 \}$ as well as global feature $\globalfeature$.
In the distance regressor, we first get the point-wise local features $\{\localfeature_{p}^{0},\localfeature_{p}^{1},\localfeature_{p}^{2},\localfeature_{p}^{3}\}$ of query point $\point$ by feature interpolation \cite{qi2017pointnet++}, and then apply a four-layer MLP to regress the point-to-point distance.
The formula of feature interpolation \cite{qi2017pointnet++} at point $\point$ is listed below:
\begin{equation}
    \localfeature_{p}^{s}=\frac{\sum_{k=1}^{3}w_{p_{k}} \localfeature_{p_{k}}^{s}}{\sum_{k=1}^{3}w_{p_{k}}}, 
    where~w_{p_{k}} = \frac{1}{||\point - \point_{k}||_2}
\end{equation}
where $\localfeature_{p}^{s}$ ($s\in\{0,1,2,3\}$) are the interpolated multi-scale local features, 
$\point_k$ ($k \in \{1,2,3\}$) are the three nearest-neighbors of query point $\point$ in the initial interpolated point cloud, 
and $||\point - \point_{k}||_2$ is the distance between $\point$ and $\point_{k}$.

\section{Additional Ablation Studies}
In this section, we conduct more ablation experiments to validate our choices of regressor input, training scheme with Gaussian noise and  location refinement strategy.
All the evaluations are done on the PU1K dataset with $4\times$ upsampling.

\paragraph{Regressor Input}
In our full model, we regress the point-to-point distance of each query point using the concatenation of coordinate $\point$, interpolated local features $\{\localfeature_{\point}^0, \localfeature_{\point}^1, \localfeature_{\point}^2, \localfeature_{\point}^3\}$ and global feature $\globalfeature$.
We here analyze the impact of each type of feature on the final performance.
We train models with either local feature or global feature, and show their performances in \Tab{regressorinput}. Both cases perform worse than our full model, which uses both features.

\begin{table}[htbp]
\small
\centering
\vspace{2pt}
\resizebox{0.78\linewidth}{!}{
\begin{tabular}{lccc}
\toprule    
 Regressor Inputs & \begin{tabular}[c]{@{}c@{}} CD $\downarrow$ \\ $10^{-3}$ \end{tabular} & \begin{tabular}[c]{@{}c@{}} HD $\downarrow$ \\ $10^{-3}$ \end{tabular} & \begin{tabular}[c]{@{}c@{}} P2F $\downarrow$ \\ $10^{-3}$ \end{tabular} \\
\midrule
$(\point, \globalfeature)$ &  0.729 & 8.439 & 2.324 \\
$(\point, \localfeature_{\point}^0, \localfeature_{\point}^1, \localfeature_{\point}^2, \localfeature_{\point}^3)$ & 0.423 & 4.011 & 1.652 \\
\hline
  \bm{$(\point, \localfeature_{\point}^0, \localfeature_{\point}^1, \localfeature_{\point}^2, \localfeature_{\point}^3, \globalfeature)$} &\textbf{0.404} & \textbf{3.732} 
  & \textbf{1.474} \\
\bottomrule
\end{tabular}}
\caption{\label{tab:regressorinput} $4\times$ comparative results on the PU1K dataset with different regressor inputs. Utilizing both local features and global feature is clearly more effective.}
\end{table}

\paragraph{Training Scheme with Gaussian Noise}
During training, we jitter the interpolated points $\point \in \interpolatedpointcloud$ with Gaussian noise to obtain query points,
which simulates the iterative optimization process of inference.
To verify the effectiveness of this training scheme, we train our P2PNet with or without Gaussian noise, while sharing the same testing process.
As shown in \Tab{gaussiannoise}, using Gaussian noise for training achieves superior performance, since it simulates the upsampling error from not just the initial interpolated point cloud but all iterations,
and also increases the smoothness and continuity of learned distance functions.

\begin{table}[htbp]
\centering
\resizebox{0.75\linewidth}{!}{
\begin{tabular}{lccc}
\toprule    
\begin{tabular}[c]{@{}c@{}}  Training \end{tabular} & \begin{tabular}[c]{@{}c@{}} CD $\downarrow$ \\ $10^{-3}$ \end{tabular} & \begin{tabular}[c]{@{}c@{}} HD $\downarrow$ \\ $10^{-3}$ \end{tabular} & \begin{tabular}[c]{@{}c@{}} P2F $\downarrow$ \\ $10^{-3}$ \end{tabular} \\
\midrule
w/o Gaussian Noise &  0.520 & 5.368 & 1.729 \\
\textbf{w Gaussian Noise} &\textbf{0.404} & \textbf{3.732} & \textbf{1.474} \\
\bottomrule
\end{tabular}}
\caption{\label{tab:gaussiannoise} $4\times$ comparative results on the PU1K dataset with or without Gaussian noise. Using Gaussian noise for training significantly improves the upsampling performance.}
\end{table}

\paragraph{Location Refinement Strategy}
After midpoint interpolation, we iteratively move the interpolated point $\point$ towards the ground truth position, guided by the estimated point-to-point distance $P2PNet(\point)$ and predefined step size $\lambda$, as formulated below:
\begin{equation}
\begin{split}
    \point^{t+1}=\point^{t}-\stepsize \nabla P2PNet(\point^{t})
    \label{eq:ourupdate}
\end{split}
\end{equation}
where $t \in [0, \iteration-1]$, $\iteration$ is the predefined iteration number.

And Chibane \etal \cite{chibane2020neural} also propose to project point by moving it the predicted distance along the normalized negative gradient, as listed below:
\begin{equation}
\begin{split}
    \point^{t+1}=\point^{t}-P2PNet(\point^{t})\frac{\nabla P2PNet(\point^{t})}{||\nabla P2PNet(\point^{t})||_2}
    \label{eq:otherupdate}
\end{split}
\end{equation}

We use \Eq{ourupdate} and \Eq{otherupdate} as the refinement strategy respectively, and we also fine-tune the iteration number $\iteration$ for \Eq{otherupdate} to achieve the best performance. As shown in \Tab{update}, our \Eq{ourupdate} is much more accurate than \Eq{otherupdate}.
While \Eq{otherupdate} requires that the estimated distance $P2PNet(\point)$ and direction $-\nabla P2PNet(\point)$  should be both accurate enough to achieve a good performance.
Our \Eq{ourupdate} only requires the accuracy of the direction. 
And once the direction is precisely predicted, the convergence of the interpolated point can be guaranteed, given sufficient iterations and a suitable step size \cite{luo2021score}.

\begin{table}[htbp]
\centering
\resizebox{0.8\linewidth}{!}{
\begin{tabular}{lccc}
\toprule    
\begin{tabular}[c]{@{}c@{}}  Refinement Strategies \end{tabular} & \begin{tabular}[c]{@{}c@{}} CD $\downarrow$ \\ $10^{-3}$ \end{tabular} & \begin{tabular}[c]{@{}c@{}} HD $\downarrow$ \\ $10^{-3}$ \end{tabular} & \begin{tabular}[c]{@{}c@{}} P2F $\downarrow$ \\ $10^{-3}$ \end{tabular} \\
\midrule
\Eq{otherupdate} & 0.872  & 8.650 & 2.541  \\
  \textbf{\Eq{ourupdate}} &\textbf{0.404} & \textbf{3.732} & \textbf{1.474} \\
\bottomrule
\end{tabular}}
\caption{\label{tab:update} $4\times$ comparative results on the PU1K dataset with different refinement strategies. Our \Eq{ourupdate} achieves significantly better performance.}
\vspace{-5pt}
\end{table}

\section{Additional Qualitative Results}
\begin{figure*}[htbp]
\vspace{-10pt}
\centering
\includegraphics[width=\textwidth]{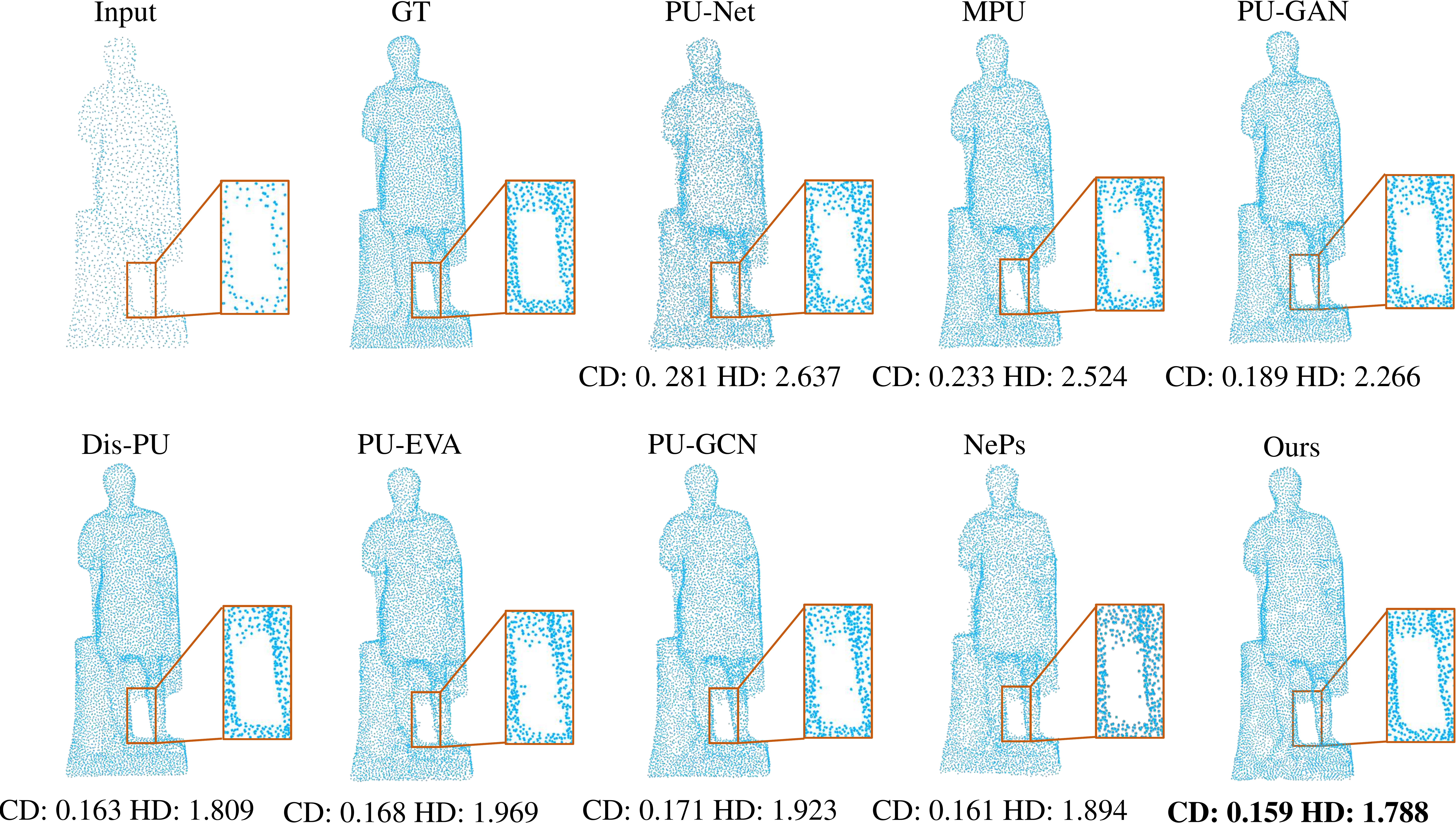}
\caption{$4\times$ upsampled results on the PU-GAN dataset. Our method produces much less outliers, more smooth surfaces and more fine-grained details. 
}
\label{fig:supp_4x}
\vspace{-20pt}
\end{figure*}

\begin{figure*}[htbp]
\vspace{-10pt}
\centering
\includegraphics[width=\textwidth]{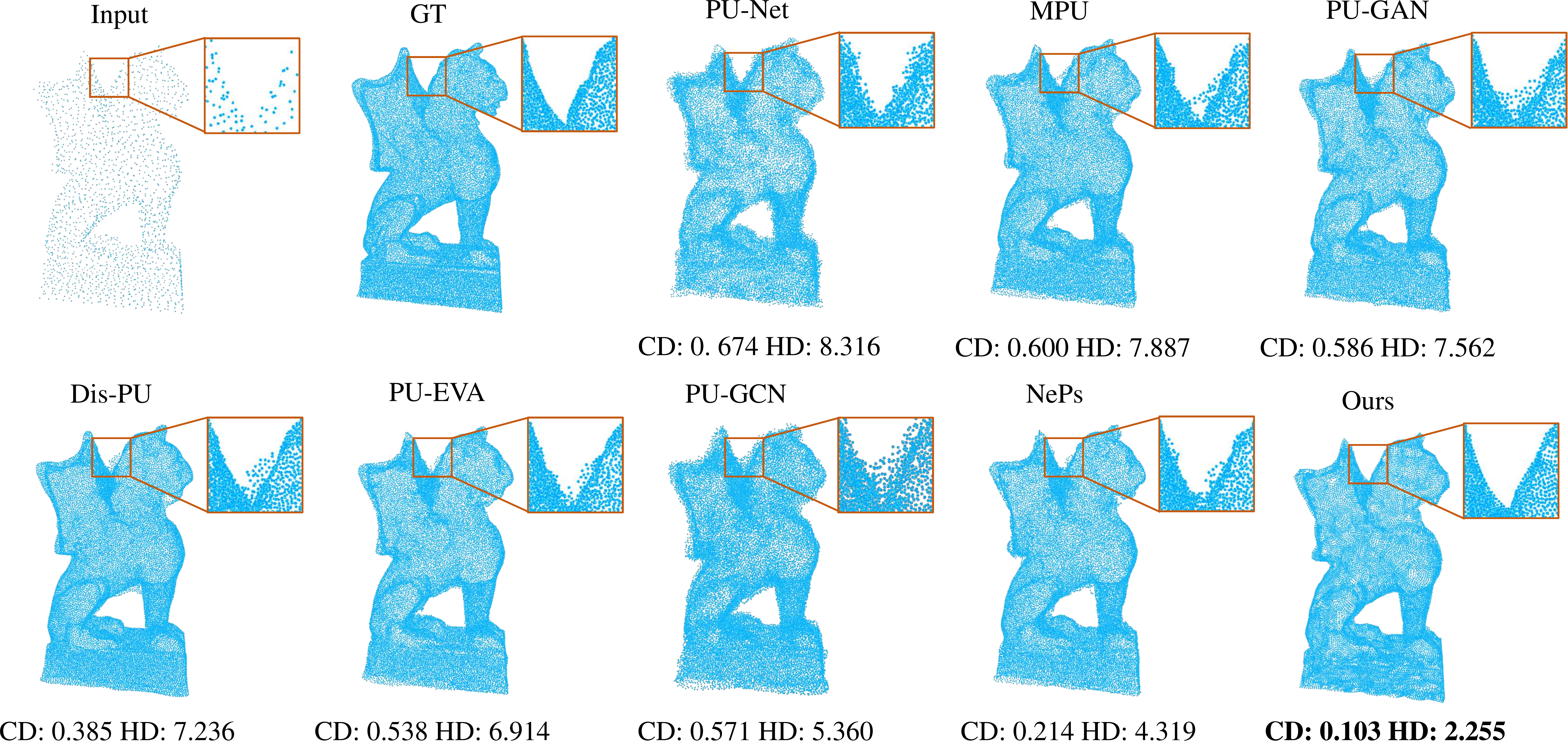}
\caption{$16\times$ upsampled results on the PU-GAN dataset. Our method produces much less outliers, more smooth surfaces and more fine-grained details. 
}
\label{fig:supp_16x}
\vspace{-20pt}
\end{figure*}

\begin{figure*}[htbp]
\vspace{-15pt}
\centering
\includegraphics[width=.9\textwidth]{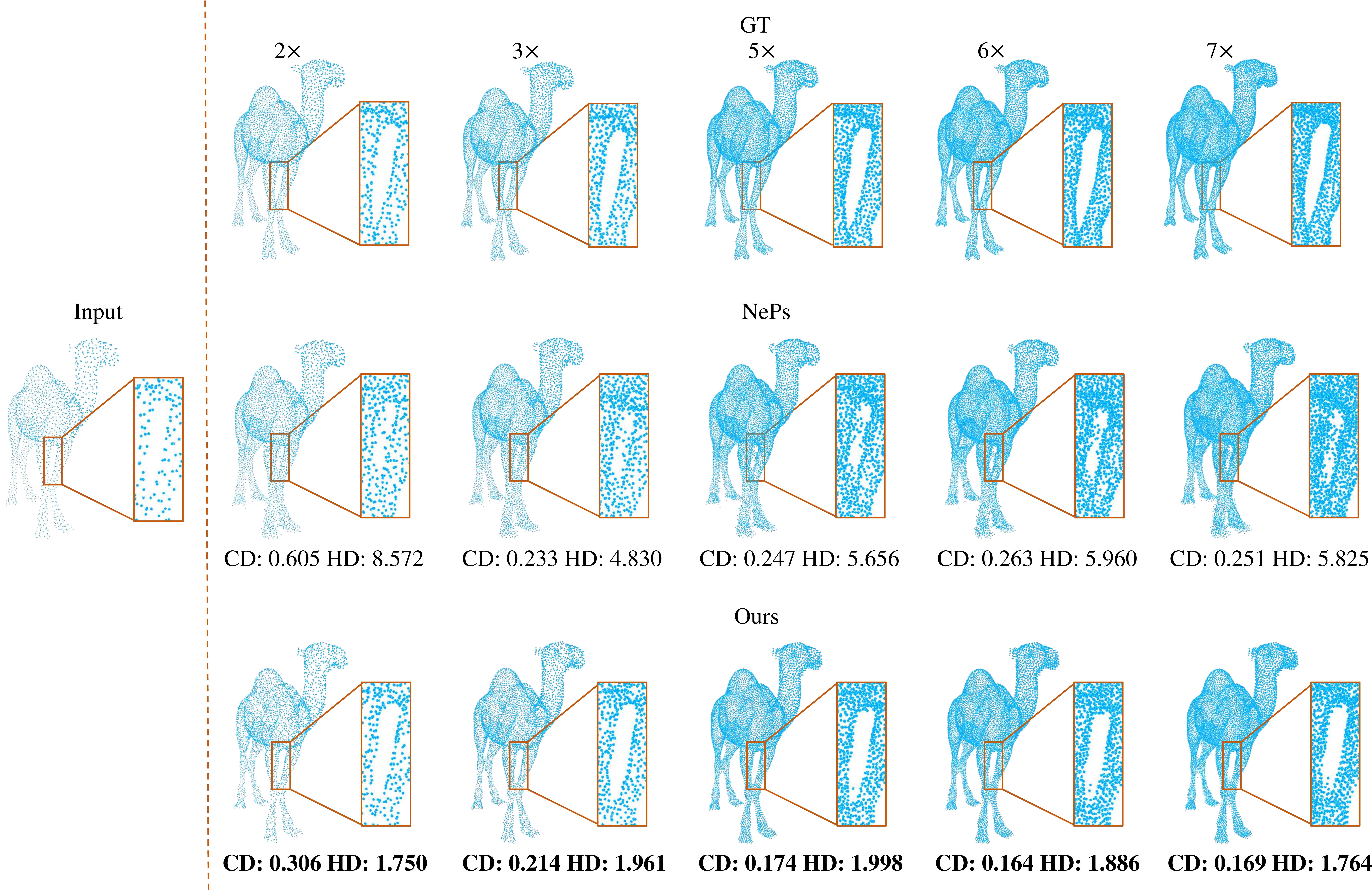}
\caption{Qualitative comparisons on the PU-GAN dataset with upsampling rate $\upsamplingrate \in \{2,3,5,6,7\}$. Note that we use the same trained model for different upsampling rates.
Our method obviously achieves superior performance across the full range of upsampling rates. }
\label{fig:supp_arbitrary}
\vspace{-20pt}
\end{figure*}

\begin{figure*}[htbp]
\vspace{-10pt}
\centering
\includegraphics[width=.9\textwidth]{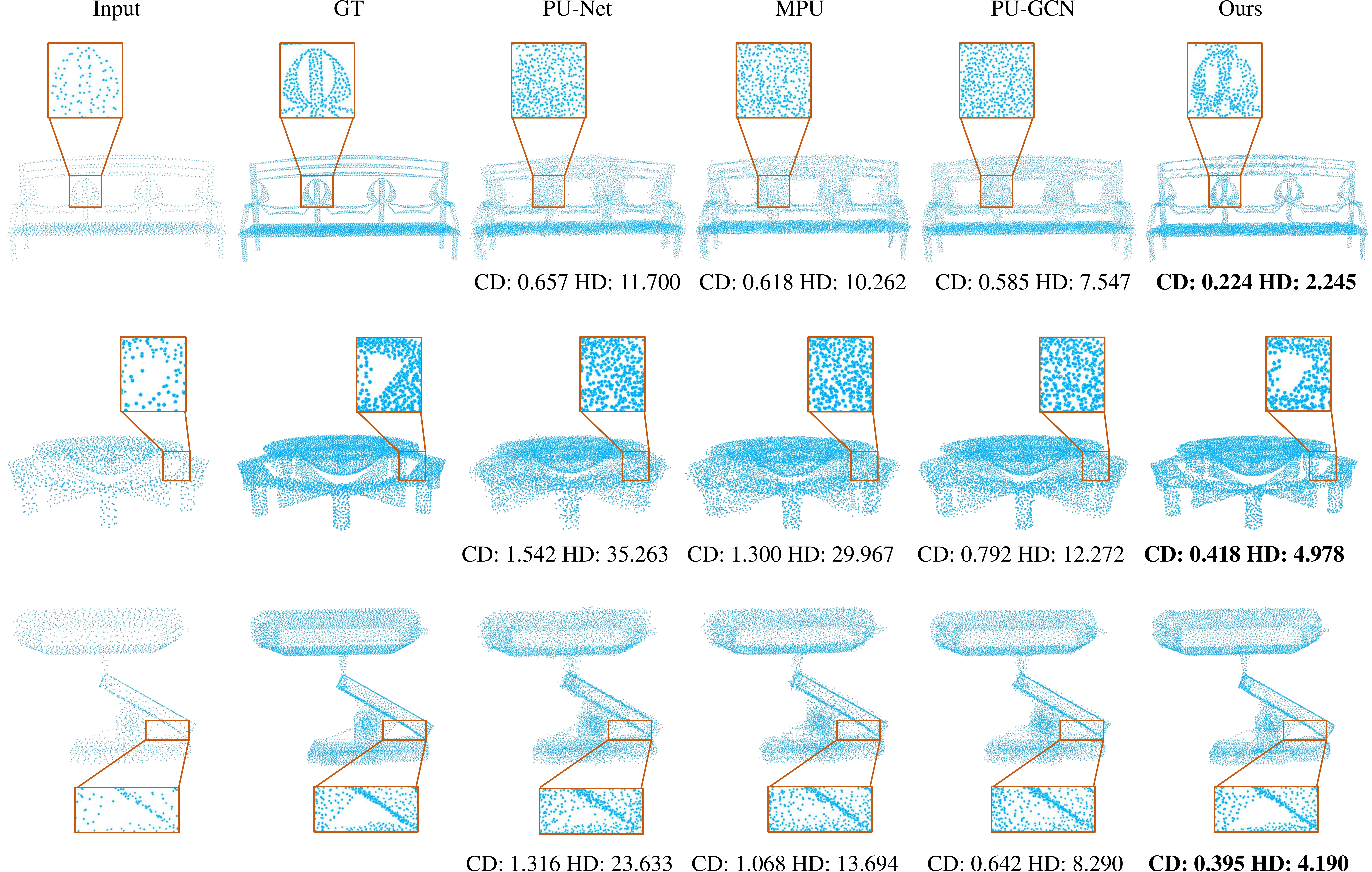}
\caption{$4\times$ qualitative comparisons on the PU1K dataset. Our results preserve more fine-grained details and produce much less outliers. }
\label{fig:supp_pu1k}
\vspace{-20pt}
\end{figure*}

\begin{figure*}[htbp]
\vspace{5pt}
\centering
\includegraphics[width=\textwidth]{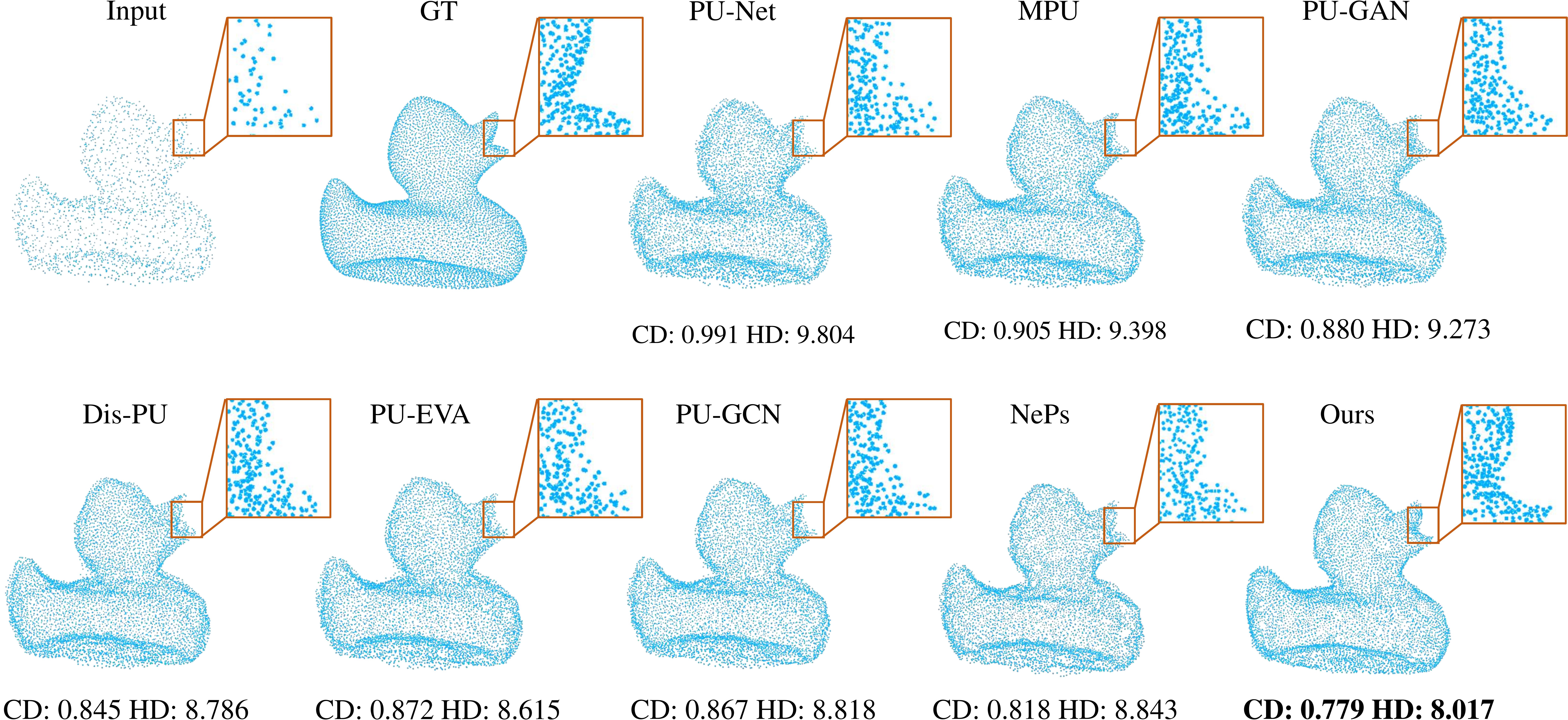}
\caption{4× qualitative results on the PU-GAN dataset with noise level $\noiselevel=0.02$. Our method generates cleaner and smoother point cloud, while preserving more details.}
\label{fig:supp_noise}
\vspace{-15pt}
\end{figure*}

\begin{figure*}[htbp]
\centering
\includegraphics[width=\textwidth]{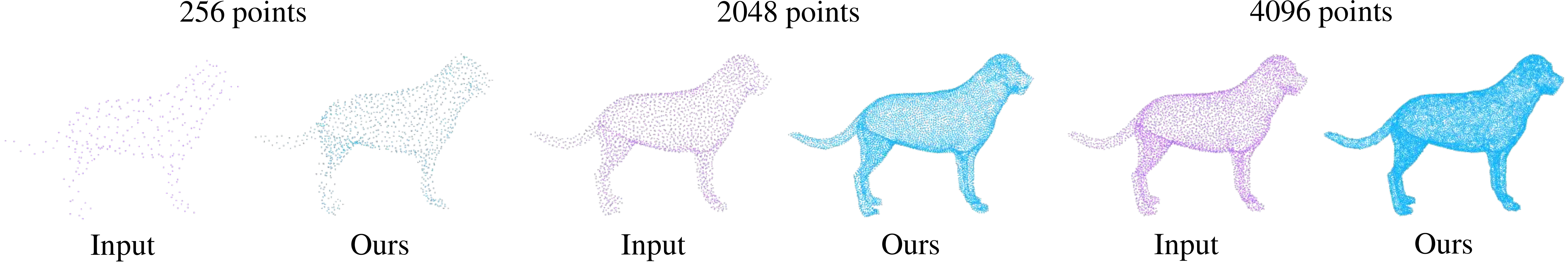}
\caption{4× upsampled results by our method with different input sizes. Our method consistently achieves high upsampling quality even when the input is extremely sparse.}
\label{fig:supp_inputsize}
\vspace{-15pt}
\end{figure*}

\begin{figure*}[t!]
\centering
\includegraphics[width=\textwidth]{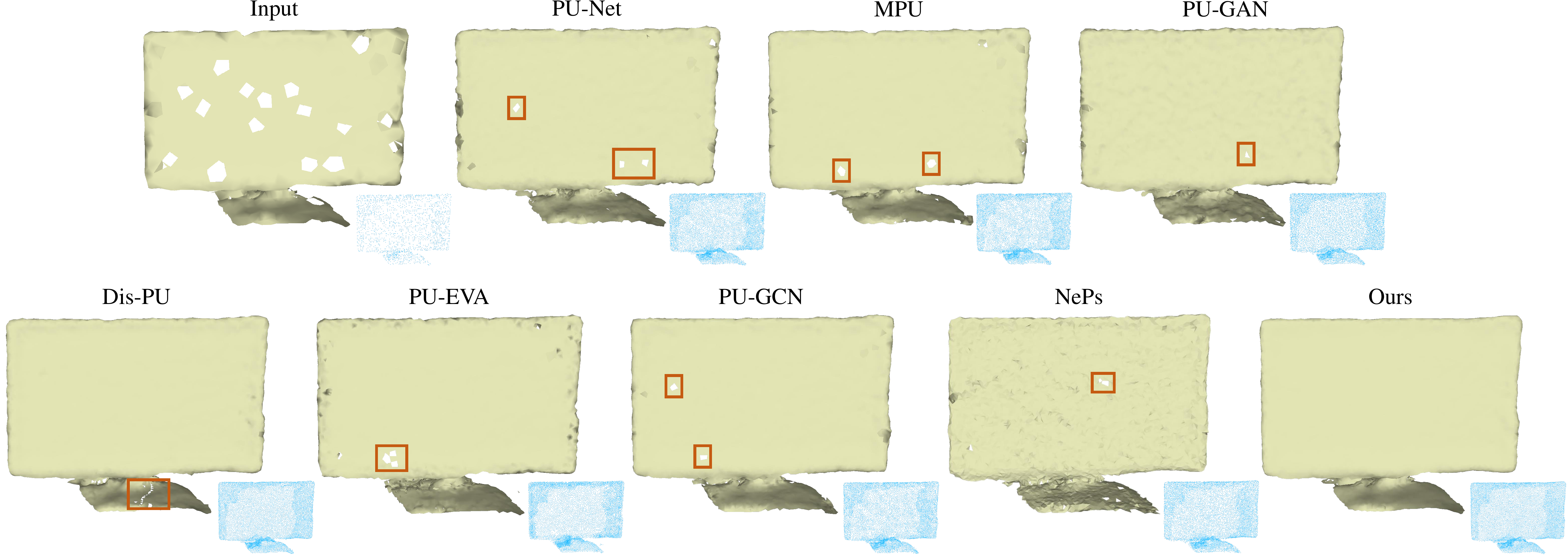}
\caption{$4\times$ upsampled results on the ScanObjectNN dataset, and the meshes are reconstructed by BallPivoting algorithm \cite{bernardini1999ball}. Our method clearly generates more complete, smooth and faithful mesh and point cloud, while other methods tend to keep the holes.}
\label{fig:supp_scanobjectnn}
\end{figure*}

\begin{figure*}[t!]
\centering
\includegraphics[width=\textwidth]{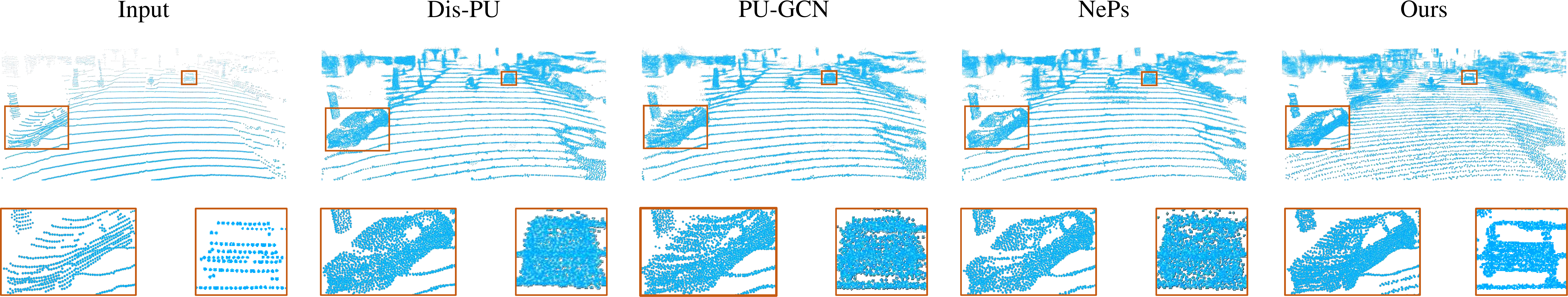}
\vspace{-10pt}
\caption{$4\times$ upsampled results on the KITTI dataset. Our result not only retains more fine-grained details but also fills the gaps between LiDAR fibers.}
\label{fig:supp_kitti}
\end{figure*}

In this section, we show some more qualitative results on the PU-GAN \cite{li2019pu}, PU1K \cite{qian2021pu}, ScanObjectNN \cite{uy2019revisiting} and KITTI \cite{geiger2013vision} datasets, which further validate that our method achieves superior upsampling quality.
Specifically, in \Fig{supp_4x} and \Fig{supp_16x}, we provide more visual comparisons with previous SOTA methods on the PU-GAN dataset, under $4\times$ and $16\times$ settings respectively.
In \Fig{supp_arbitrary}, we visualize the upsampled results with various upsampling rates after one-time training.
In \Fig{supp_pu1k}, we present qualitative comparisons on the PU1K dataset with $4\times$ setting.
In \Fig{supp_noise} and \Fig{supp_inputsize}, we employ added noise and different input sizes to verify the robustness of our method.
In \Fig{supp_scanobjectnn} and \Fig{supp_kitti}, we display more qualitative results on the real-scanned inputs.
Consistent with the main paper, we use CD and HD to represent Chamfer distance and Hausdorff distance respectively, while their units are both $10^{-3}$.

